\documentclass{article}

\usepackage[preprint]{nips_2018}

\usepackage[utf8]{inputenc} 
\usepackage[T1]{fontenc}    
\usepackage{hyperref}       
\usepackage{url}            
\usepackage{booktabs}       
\usepackage{amsfonts}       
\usepackage{nicefrac}       
\usepackage{microtype}      
\usepackage{amsmath}
\usepackage{subcaption}
\usepackage{graphicx}
\usepackage{multirow}
\usepackage{times}
\usepackage{epsfig}
\usepackage{amssymb}
\usepackage{colortbl}
\usepackage{afterpage}
\usepackage{siunitx}

\definecolor{mygray}{gray}{.9}
\definecolor{myhoneydew}{rgb}{0.94,1.0,0.94}
\definecolor{mysky}{rgb}{0.53,0.81,0.98}
\definecolor{myazure}{rgb}{0.94,1.0,1.0}

\newcommand{\etal}{\emph{et al.}~}

\title{\Large{Geometry-Consistent Generative Adversarial Networks for One-Sided Unsupervised  Domain Mapping}}

\author{ 
Huan Fu\thanks{equal contribution} \ $^1$\ \ \ \ \ \ \  Mingming Gong$^{\ast}$ $^{2,3}$\ \  \\
\\
\textbf{Chaohui  Wang}${^{4}}$\ \ \ \  \textbf{Kayhan  Batmanghelich}$^{2, 3}$\ \ \ \   \textbf{Kun Zhang}${^{3}}$\ \ \ \    \textbf{Dacheng Tao}$^1$ \\
\\
{$^1$UBTECH Sydney AI Centre, SIT, FEIT, The University of Sydney, Australia}\\ 
{$^2$Department of Biomedical Informatics, University of Pittsburgh}\\ 
{$^3$Department of Philosophy, Carnegie Mellon University}\\ 
{$^4$Laboratoire d'Informatique Gaspard Monge, Universit\'{e} Paris-Est}\\
{\tt\footnotesize \{hufu6371@uni., dacheng.tao@\}sydney.edu.au \  \{mig73, kayhan\}@pitt.edu} \\
{\tt\footnotesize chaohui.wang@u-pem.fr \ kunz1@cmu.edu } 
}

\begin{document}

\maketitle

\begin{abstract}
Unsupervised domain mapping aims to learn a function to translate domain $\mathcal{X}$ to $\mathcal{Y}$ by a function $G_{XY}$ in the absence of paired examples. Finding the optimal $G_{XY}$ without paired data is an ill-posed problem, so appropriate constraints are required to obtain reasonable solutions. One of the most prominent constraints is cycle consistency, which enforces the translated image by $G_{XY}$ to be translated back to the input image by an inverse mapping $G_{YX}$. While cycle consistency requires the simultaneous training of $G_{XY}$ and $G_{YX}$, recent studies have shown that one-sided domain mapping can be achieved by preserving pairwise distances between images. Although cycle consistency and distance preservation successfully constrain the solution space, they overlook the special properties of images that simple geometric transformations do not change the image's semantic structure. Based on this special property, we develop a geometry-consistent generative adversarial network (GcGAN), which enables one-sided unsupervised domain mapping. GcGAN takes the original image and its counterpart image transformed by a predefined geometric transformation as inputs and generates two images in the new domain coupled with the corresponding geometry-consistency constraint. The geometry-consistency constraint reduces the space of possible solutions while keep the correct solutions in the search space. Quantitative and qualitative comparisons with the baseline (GAN alone) and the state-of-the-art methods including CycleGAN \cite{zhu2017unpaired} and DistanceGAN \cite{benaim2017one} demonstrate the effectiveness of our method.
\end{abstract}

\section{Introduction}
Domain mapping or image-to-image translation, which targets at translating an image from one domain to another, has been intensively investigated over the past few years.  Let $X\in\mathcal{X}$ denote a random variable representing source domain images and $Y\in\mathcal{Y}$ represent target domain images. 
According to whether we have access to a paired sample $\{(x_{i}, y_{i})\}_{i=1}^{N}$, domain mapping can be studied in a supervised or unsupervised manner. While several works have successfully produced high-quality translations by focusing on supervised domain mapping with constraints provided by cross-domain image pairs \cite{pathak2016context, isola2017image,wang2017high, Wang_2018_ECCV}, the progress of unsupervised domain mapping is relatively slow. 
Unluckily, obtaining paired training examples is expensive and even infeasible in some situations. For example, if we want to learn translators between Monet's paintings and Photographs, how can we collect sufficient well-defined (\emph{Monet's painting, photograph}) pairs for model training? By contrast, collecting unpaired sets is often convenient since infinite images are available online. From this viewpoint, unsupervised domain mapping has great potential for real-world applications in the long term. 

\begin{figure*}[t]
\begin{center}
\begin{subfigure}{0.98\textwidth}
  \begin{center}
  \scalebox{1}[1]{\includegraphics[scale=0.43]{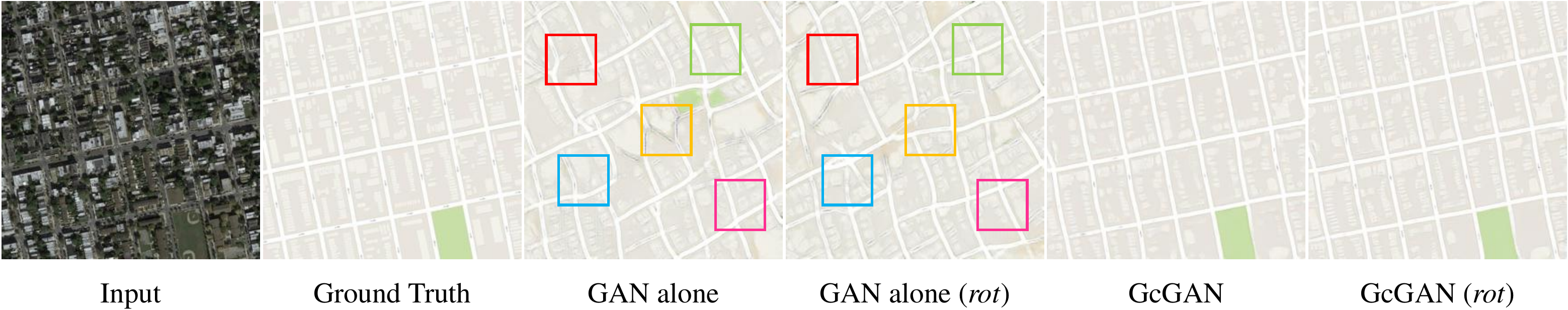}}
  \end{center}
\end{subfigure}%
\captionsetup{font={small}}
\caption{\textbf{Geometry consistency.} The original input image is denoted by $x$, and the predefined function $f(\cdot)$ is a $\ang{90}$ clockwise rotation (\emph{rot}). GAN alone: $G_{XY}^{1}(x)$. GAN alone (\emph{rot}): $f^{-1}(G_{\tilde{X}\tilde{Y}}^{1}(f(x)))$. GcGAN: $G_{XY}^{2}(x)$. GcGAN (\emph{rot}): $f^{-1}(G_{\tilde{X}\tilde{Y}}^{2}(f(x))$. It can be seen that GAN alone produces geometrically-inconsistent output images, indicating that the learned $G_{XY}$ and $G_{\tilde{X}\tilde{Y}}$ are far away from the correct mapping functions. By enforcing geometry consistency, our method results in more sensible domain mapping. $GcGAN =  GAN \ alone + geometry \ consistency$.}
\label{fig:rotation}
\end{center}
\vspace{-0.55cm}
\end{figure*}

In unsupervised domain mapping, from a probabilistic modeling perspective, our goal is to model the joint distribution $P_{XY}$ given samples drawn from the marginal distributions $P_X$ and $P_Y$ in individual domains. Since the two marginal distributions can be inferred from an infinite set of possible joint distributions, it is difficult to guarantee that an individual input $x \in X$ and the output $G_{XY}(x)$ are paired up in a meaningful way without additional assumptions or constraints. 

To address this problem, recent approaches have exploited the cycle-consistency assumption, {\it i.e}., a mapping $G_{XY}$ and its inverse mapping $G_{YX}$ should be bijections \cite{zhu2017unpaired, kim2017learning, yi2017dualgan}. Specifically, when feeding an example $x \in X$ into the networks $G_{XY} \circ G_{YX}: X \to Y \to X$, the output should be a reconstruction of $x$ and vise versa for $y$, {\it i.e}., $G_{YX}(G_{XY}(x)) \approx x$ and $G_{XY}(G_{YX}(y)) \approx y$. Further, DistanceGAN \cite{benaim2017one} showed that maintaining the distances between images within domains allows one-sided unsupervised domain mapping rather than simultaneously learning both $G_{XY}$ and $G_{YX}$. 



Existing constraints overlook the special properties of images that simple geometric transformations (global geometric transformations without shape deformation) do not change the image's semantic structure. Here, semantic structure refers to the information that distinguishes different object/staff classes, which can be easily perceived by humans regardless of trivial geometric transformations such as rotation. Based on this property, we develop a geometry-consistency constraint, which helps in reducing the search space of possible solutions while still keeping the correct set of solutions under consideration, and results in a geometry-consistent generative adversarial network (GcGAN) for unsupervised domain mapping. 

Our geometry-consistency constraint is motivated by the fact that a given geometric transformation $f(\cdot)$ between the input images should be preserved by related translators $G_{XY}$ and $G_{\tilde{X}\tilde{Y}}$,  if $\tilde{\mathcal{X}}$ and $\tilde{\mathcal{Y}}$ are the domains obtained by applying $f(\cdot)$ on the examples of $X$ and $Y$, respectively. Mathematically, given a random example $x$ from the source domain $\mathcal{X}$ and a predefined geometric transformation function $f(\cdot)$, geometry consistency can be expressed as $f(G_{XY}(x)) \approx G_{\tilde{X}\tilde{Y}}(f(x))$ and $f^{-1}(G_{\tilde{X}\tilde{Y}}(f(x))) \approx G_{XY}(x)$, where $f^{-1}(\cdot)$ is the inverse function of $f(\cdot)$. Because it is unlikely that $G_{XY}$ and $G_{\tilde{X}\tilde{Y}}$ always fail in the same location,   $G_{XY}$ and $G_{\tilde{X}\tilde{Y}}$ co-regularize each other by the geometry-consistency constraint and thus correct each others' failures in local regions of their respective translations (see Figure~\ref{fig:rotation} for an illustrative example). Our geometry-consistency constraint allows one-sided unsupervised domain mapping, {\it i.e}., $G_{XY}$ can be trained independently from $G_{YX}$. In this paper, we employ two simple but representative geometric transformations as examples, {\it i.e}., vertical flipping (\emph{vf}) and 90 degrees clockwise rotation (\emph{rot}), to illustrate geometry consistency. Quantitative and qualitative comparisons with the baseline (GAN alone) and the state-of-the-art methods including CycleGAN \cite{zhu2017unpaired} and DistanceGAN \cite{benaim2017one} demonstrate the effectiveness of our model in generating realistic images.
\begin{figure*}[t]
\begin{center}
\begin{subfigure}{1.0\textwidth}
  \begin{center}
  \includegraphics[scale=0.45]{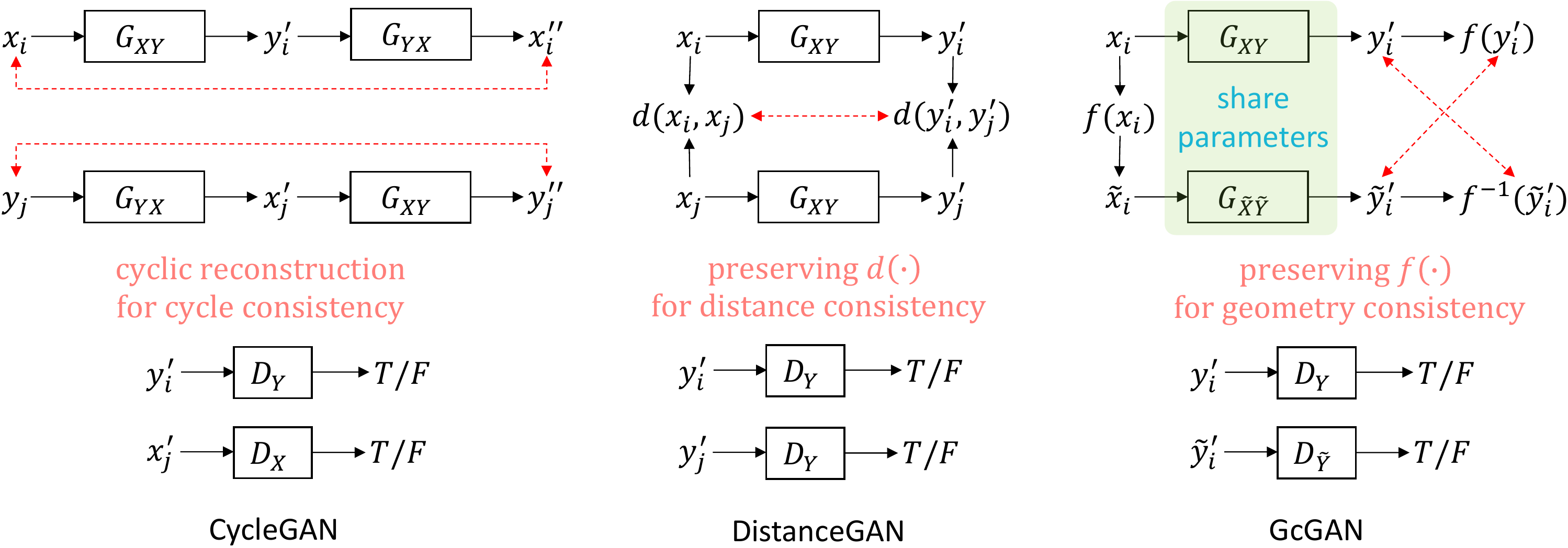}
  \end{center}
\end{subfigure}%
\captionsetup{font={small}}
\caption{\textbf{An illustration of the differences between CycleGAN \cite{zhu2017unpaired}, DistanceGAN \cite{benaim2017one}, and our GcGAN}. $x$ and $y$ are random examples from domain $\mathcal{X}$ and $\mathcal{Y}$, respectively. $d(x_{i}, x_{j})$ is the distance between images $x_i$ and $x_j$. $f(\cdot)$ is a predefined geometric transformation function for images, which satisfies $f^{-1}(f(x)) = f(f^{-1}(x)) = x$.  $G_{XY}$ and $G_{\tilde{X}\tilde{Y}}$  are the generators (or translators) which target the domain translation tasks from $\mathcal{X}$ to $\mathcal{Y}$ and $\tilde{\mathcal{X}}$ to $\tilde{\mathcal{Y}}$, where $\tilde{\mathcal{X}}$ and $\tilde{\mathcal{Y}}$ are two domains obtained by applying $f(\cdot)$ on all the images in $\mathcal{X}$ and $\mathcal{Y}$, respectively. $D_{Y}$ is an adversarial discriminator in domain $\mathcal{Y}$. The red dotted lines denote the unsupervised constraints with respect to cycle consistency ($x \approx G_{YX}(G_{XY}(x))$), distance consistency ($x \approx G_{YX}(G_{XY}(x))$), and our geometry consistency ($f(G_{XY}(x)) \approx G_{\tilde{X}\tilde{Y}}(f(x))$), respectively. }
\label{fig:gc-gan}
\end{center}
\vspace{-0.4cm}
\end{figure*}
\section{Related Work}
\noindent \textbf{Generative Adversarial Networks.} 
Generative adversarial networks (GANs) \cite{goodfellow2014generative, oord2016conditional, denton2015deep, radford2015unsupervised, salimans2016improved, arjovsky2017wasserstein} learn two networks, {\it i.e}., a generator and a discriminator, in a staged zero-sum game fashion to generate images from inputs. Many applications and computer vision tasks have recently been developed based on deep convolutional GANs (DCGANs), such as image inpainting, text to image synthesis, style transfer, and domain adaptation \cite{bousmalis2016domain, zhang2017stackgan, pathak2016context, reed2016generative, ledig2017photo, wang2018mix, chen2018stereoscopic, shen2018neural, hoffman2017cycada, sheng2018avatar, zhang2018separating, johnson2016perceptual, royer2017xgan}. The key components enabling GANs is the proposed adversarial constraint, which enforces the generated images to be indistinguishable from real images. Our formulation also benefits from an adversarial constraint to learn translators between two individual domains.
\newline

\noindent \textbf{Domain Mapping.} 
Many well-known computer vision tasks, such as scene parsing and image colorization, follow similar settings to domain mapping or image-to-image translation. Specific to recent adversarial domain mapping, this problem has been studied in a supervised or unsupervised manner with respect to paired or unpaired inputs.
 
There are a variety of literatures \cite{pathak2016context, ledig2017photo, isola2017image,wang2017high, ulyanov2016texture, Wang_2018_ECCV, isokane2018probabilistic, lin2018conditional, azadi2018multi, chen2017photographic} on supervised domain mapping. One representative example is conditional GAN \cite{isola2017image}, which learns the discriminator to distinguish $(x, y)$ and $(x, G_{XY}(x))$ instead of $y$ and $G_{XY}(x)$, where $(x, y)$ is a meaningful pair across domains.
Further, Wang \etal \cite{wang2017high} showed that conditional GANs can be used to generate high-resolution images with a novel feature matching loss, as well as multi-scale generator and discriminator architectures.
While there has been significant progress in supervised domain mapping, many real-word applications can not provide aligned images across domains because data preparation is expensive. Thus, different constraints and frameworks have been proposed for image-to-image translation in the absence of training pairs, {\it i.e}., unsupervised domain mapping.

In unsupervised domain mapping, only unaligned examples in individual domains are provided, making the task more practical but more difficult. Unpaired domain mapping has a long history, and some successes in adversarial networks have recently been presented \cite{liu2016coupled, zhu2017unpaired, benaim2017one, liu2017unsupervised, ma2018gan, liu2018unified, benaim2018one, StarGAN2018}. For example, Liu and Tuzel \cite{liu2016coupled} introduced coupled GAN (CoGAN) to learn cross-domain representations by enforcing a weight-sharing constraint. 
Subsequently, CycleGAN \cite{zhu2017unpaired}, DiscoGAN \cite{kim2017learning}, and DualGAN \cite{yi2017dualgan} enforced that translators $G_{XY}$ and $G_{YX}$ should be bijections. Thus, jointly learning $G_{XY}$ and $G_{YX}$ by enforcing cycle consistency can help to produce convincing mappings. Since then, many constraints and assumptions have been proposed to improve cycle consistency \cite{chang2018pairedcyclegan, gokaslan2018improving, huang2018multimodal, lee2018diverse, li2018unsupervised, StarGAN2018, anoosheh2018combogan, zhu2017toward, gonzalez2018image, mejjati2018unsupervised, liu2017unsupervised, liang2017generative, almahairi2018augmented}. Recently, Benaim and Wolf \cite{benaim2017one} reported that maintaining the distances between samples within domains allows one-sided unsupervised domain mapping. GcGAN is also a one-sided framework coupled with our geometry-consistency constraint, and produces competitive and even better translations than the two-sided CycleGAN in various applications.

\section{Preliminaries}
Let $\mathcal{X}$ and $\mathcal{Y}$ be two domains with unpaired training examples $\{x_{i}\}_{i=1}^{N}$ and $\{y_{j}\}_{j=1}^{M}$, where $x_{i}$ and $y_{j}$ are drawn from the marginal distributions $P_{X}$ and $P_{Y}$, where $X$ and $Y$ are two random variables associated with $\mathcal{X}$ and $\mathcal{Y}$, respectively. In the paper, we exploit style transfer without undesirable semantic distortions in unsupervised domain mapping, and have two goals.
First, we need to learn a mapping $G_{XY}$ such that $G_{XY}(X)$ has the same distribution as $Y$, {\it i.e}., $P_{G_{XY}(X)} \approx P_{Y}$. Second, the learned mapping function only changes the image style without distorting the semantic structures.

While many works have modeled the invertibility between $G_{XY}$ and $G_{YX}$ for convincing mappings since the success of CycleGAN, here we propose to enforce geometry consistency as a constraint that allows one-sided domain mapping, {\it i.e}., learning $G_{XY}$ without simultaneously learning $G_{YX}$. Let $f(\cdot)$ be a predefined geometric transformation. We can obtain two extra domains $\tilde{\mathcal{X}}$ and $\tilde{\mathcal{Y}}$ with examples $\{\tilde{x}_i\}_{i=1}^N$ and $\{\tilde{y}_j\}_{j=1}^M$ by applying $f(\cdot)$ on $X$ and $Y$, respectively. We learn an additional image-to-image translator $G_{\tilde{X}\tilde{Y}}: \tilde{X} \to \tilde{Y}$ while learning $G_{XY}: X \to Y$, and introduce our geometry-consistency constraint based on the predefined transformation such that the two networks can regularize each other. Our framework enforces that $G_{XY}(x)$ and $G_{\tilde{X}\tilde{Y}}(\tilde{x})$ should keep the same geometric transformation with the one between $x$ and $\tilde{x}$, {\it i.e}., $f(G_{XY}(x)) \approx G_{\tilde{X}\tilde{Y}}(\tilde{x})$, where $\tilde{x} = f(x)$. We denote the two adversarial discriminators as $D_{Y}$ and $D_{\tilde{Y}}$ with respect to domains $\mathcal{Y}$ and $\tilde{\mathcal{Y}}$, respectively. 

\section{Proposed Method}
We present our geometry-consistency constraint and GcGAN beginning with a review of the cycle-consistency constraint and the distance constraint. An illustration of the main differences between these constraints is shown in Figure~\ref{fig:gc-gan}.

\subsection{Unsupervised Constraints}
\noindent \textbf{Cycle-consistency constraint.} Following the cycle-consistency assumption \cite{kim2017learning, zhu2017unpaired, yi2017dualgan}, through the translators $G_{XY} \circ G_{YX}: X \to Y \to X$ and $G_{YX} \circ G_{XY}: Y \to X \to Y$, the examples  $x$ and $y$ in domain $\mathcal{X}$ and $\mathcal{Y}$ should recover the original images, {\it i.e}., $x \approx G_{YX}(G_{XY}(x))$ and $y \approx G_{XY}(G_{YX}(y))$. Cycle consistency is implemented by a bidirectional reconstruction process that requires $G_{XY}$ and $G_{YX}$ to be jointly learned, as shown in Figure \ref{fig:gc-gan} (CycleGAN). The cycle consistency loss $\mathcal{L}_{cyc}(G_{XY}, G_{YX}, X, Y)$ takes the form as:
\begin{equation}
\begin{split}
\mathcal{L}_{cyc}(G_{XY}, G_{YX}, X, Y) & = \mathbb{E}_{x \sim P_{X}}[\| G_{YX}(G_{XY}(x)) - x \|_{1}] \\
& \quad + \mathbb{E}_{y \sim P_{Y}}[\| G_{XY}(G_{YX}(y)) - y \|_{1}].
 \end{split}
 \label{eq:cycle}
\end{equation}
 \vspace{-0.5cm}
\newline

\noindent \textbf{Distance constraint.} The assumption behind the distance constraint is that the distance between two examples $x_{i}$ and $x_{j}$ in domain $X$ should be preserved after mapping to domain $Y$, {\it i.e}., $d(x_i, x_j) \approx a\cdot d(G_{XY}(x_i), G_{XY}(x_j))+b$, where $d(\cdot)$ is a predefined function to measure the distance between two examples and $a$ and $b$ are the linear coefficient and bias. In DistanceGAN \cite{benaim2017one}, the distance consistency loss $\mathcal{L}_{dis}(G_{XY}, X, Y)$ is the exception to the absolute differences between distances:
\begin{equation}
\begin{split}
\mathcal{L}_{dis}(G_{XY}, X, Y) & = \mathbb{E}_{x_{i}, x_{j} \sim P_{X}}[| \phi(x_{i}, x_{j}) - \psi(x_{i}, x_{j}) |], \\
\phi(x_{i}, x_{j}) &= \frac{1}{\sigma_{X}}(\|x_{i} - x_{j}\|_{1} - \mu_{X}), \\
\psi(x_{i}, x_{j}) &= \frac{1}{\sigma_{Y}}(\|G_{XY}(x_{i}) - G_{XY}(x_{j})\|_{1} - \mu_{Y}),
 \end{split}
 \label{eq:dis}
\end{equation}
where $\mu_{X}$, $\mu_{Y}$ ($\sigma_{X}$, $\sigma_{Y}$) are the means (standard deviations) of distances of all the possible pairs of $(x_i, x_j)$ within domain $\mathcal{X}$ and $(y_i, y_j)$ within domain $\mathcal{Y}$, respectively, and are precomputed. Distance preservation makes one-sided unsupervised domain mapping possible.

\begin{figure*}[t]
\begin{center}
\begin{subfigure}{0.98\textwidth}
  \begin{center}
  \scalebox{1}[1]{\includegraphics[scale=0.8]{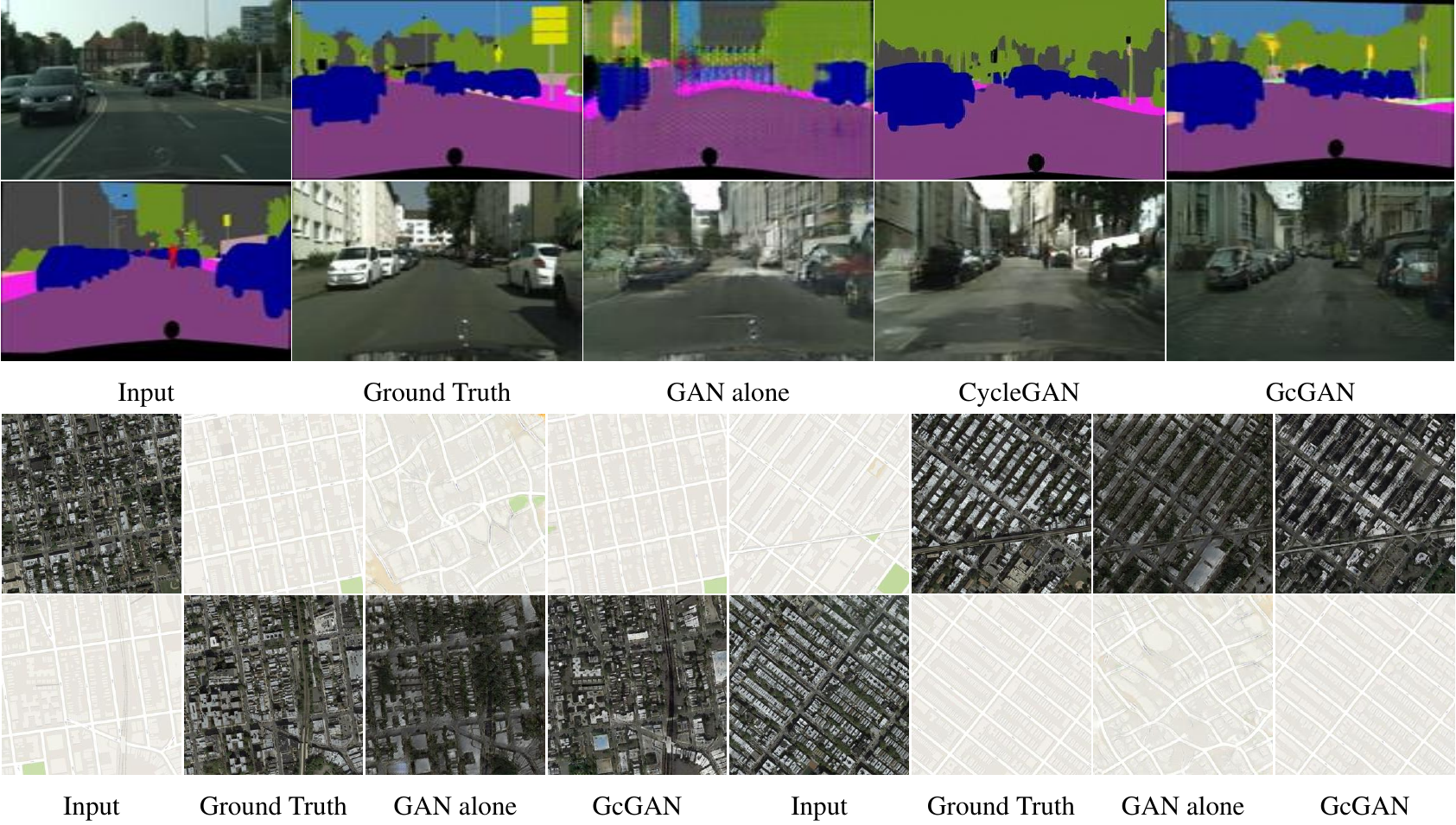}}
  \end{center}
\end{subfigure}%
\captionsetup{font={small}}
\caption{Qualitative comparison on Cityscapes (Parsing $\rightleftharpoons$ Image) and Google Maps (Map $\rightleftharpoons$ Aerial photo). GAN alone suffers from mode collapse. Translated images by GcGAN contain more details. $GcGAN =  GAN \ alone + geometry \ consistency$.}
\label{fig:cityscape}
\end{center}
\vspace{-0.4cm}
\end{figure*}

\subsection{Geometry-consistent Generative Adversarial Networks }
\noindent \textbf{Adversarial constraint.} Taking $G_{XY}$ as an example, an adversarial loss $\mathcal{L}_{gan}(G_{XY}, D_{Y}, X, Y)$ \cite{goodfellow2014generative} enforces $G_{XY}$ and $D_{Y}$ to simultaneously optimize each other in an minimax game, {\it i.e}., $\text{min}_{G_{XY}} \text{max}_{D_{Y}}\mathcal{L}_{gan}(G_{XY}, D_{Y}, X, Y)$. In other words, $D_{Y}$ aims to distinguish real examples $\{ y \}$ from translated samples $\{ G_{XY}(x) \}$. By contrast, $G_{XY}$ aims to fool $D_{Y}$ so that $D_{Y}$ can label a fake example $y' = G_{XY}(x)$ as a sample satisfying $y' \sim P_{Y}$. The objective can be expressed as: 
\begin{equation}
\begin{split}
\mathcal{L}_{gan}(G_{XY}, D_{Y}, X, Y)  & = \mathbb{E}_{y \sim P_{Y}}[\log D_{Y}(y)] \\
& \quad + \mathbb{E}_{x \sim P_{X}}[\log(1 - D_{Y}(G_{XY}(x)))].
 \end{split}
 \label{eq:gan}
\end{equation}
In the transformed domains $\tilde{\mathcal{X}}$ and $\tilde{\mathcal{Y}}$, we employ the adversarial loss $\mathcal{L}_{gan}(G_{\tilde{X}\tilde{Y}}, D_{\tilde{Y}}, \tilde{X}, \tilde{Y})$ that has the same form to $\mathcal{L}_{gan}(G_{XY}, D_{Y}, X, Y)$.
\newline

\noindent \textbf{Geometry-consistency constraint.} As shown in Figure~\ref{fig:gc-gan} (GcGAN), given a predefined geometric transformation function $f(\cdot)$, we feed the images $x \in X$ and $\tilde{x} = f(x)$ into the translators $G_{XY}$ and $G_{\tilde{X}\tilde{Y}}$, respectively. Following our geometry-consistency constraint, the outputs $y' = G_{XY}(x)$ and $\tilde{y}' = G_{\tilde{X}\tilde{Y}}(\tilde{x})$ should also satisfy $\tilde{y}' \approx f(y')$ like $x$ and $\tilde{x}$. Considering both $f(\cdot)$ and the inverse geometric transformation function $f^{-1}(\cdot)$, our complete geometry consistency loss $\mathcal{L}_{{geo}}(G_{XY}, G_{\tilde{X}\tilde{Y}}, X, Y) $ has the following form:
\begin{equation}
\begin{split}
\mathcal{L}_{{geo}}(G_{XY}, G_{\tilde{X}\tilde{Y}}, X, Y)  & = \mathbb{E}_{x \sim P_{X}}[\| G_{XY}(x) - f^{-1}(G_{\tilde{X}\tilde{Y}}(f(x))) \|_{1}]  \\
& \quad + \mathbb{E}_{x \sim P_X}[\| G_{\tilde{X}\tilde{Y}}(f(x)) - f(G_{XY}(x)) \|_{1}].
 \end{split}
 \label{eq:geo}
\end{equation}
This geometry-consistency loss can be seen as a reconstruction loss that relies on the predefined geometric transformation function $f(\cdot)$. 
In this paper, we only take two common geometric transformations as examples, namely vertical flipping (\emph{vf}) and $\ang{90}$ clockwise rotation (\emph{rot}), to demonstrate the effectiveness of our geometry-consistency constraint. Note that, $G_{XY}$ and $G_{\tilde{X}\tilde{Y}}$ have the same architecture and share all the parameters.
\newline

\noindent \textbf{Full objective.} By combining our geometry-consistency constraint with the standard adversarial constraint, a remarkable one-sided unsupervised domain mapping can be targeted. The full objective for our GcGAN $\mathcal{L}_{GcGAN}(G_{XY}, G_{\tilde{X}\tilde{Y}}, D_{Y}, D_{\tilde{Y}}, X, Y)$ will be:
\begin{equation}
\begin{split}
\mathcal{L}_{GcGAN}(G_{XY}, G_{\tilde{X}\tilde{Y}}, D_{Y}, D_{\tilde{Y}}, X, Y)  & = \mathcal{L}_{gan}(G_{XY}, D_Y, X, Y) \\
& \quad +  \mathcal{L}_{gan}(G_{\tilde{X}\tilde{Y}}, D_{\tilde{Y}}, X, Y) \\
& \quad + \lambda\mathcal{L}_{geo}(G_{XY}, G_{\tilde{X}\tilde{Y}}, X, Y),
 \end{split}
 \label{eq:net_a}
\end{equation}
where $\lambda$ ($\lambda = 20.0$ in all the experiments) is a trade-off hyperparameter to weight the contribution of $\mathcal{L}_{gan}$ and $\mathcal{L}_{geo}$ during the model training. Because that we do not make great effects to choose $\lambda$, heavily tuning $\lambda$ may give preferable results to specific translation tasks.
\newline


\noindent \textbf{Network architecture.}
The full framework of our GcGAN is illustrated in Figure~\ref{fig:gc-gan}. Our experimental settings, network architectures, and learning strategies follow CycleGAN. We employ the same discriminator and generator as CycleGAN depending on the specific tasks. Specifically, the generator is a standard encoder-decoder, where the encoder contains two convolutional layers with stride 2 and several residual blocks \cite{he2016deep} (6 / 9 blocks with respect to $128 \times 128$ / $256 \times 256$ of input resolution), and the decoder contains two deconvolutional layers also with stride 2. The discriminator distinguishes images at the patch level following PatchGANs \cite{isola2017image, li2016precomputed}. Like CycleGAN, we also use an identity mapping loss \cite{taigman2016unsupervised} in all of our experiments (except SVHN $\to$ MNIST), including our baseline (GAN alone). For other details, we use LeakyReLU as nonlinearity for the discriminators and instance normalization \cite{ulyanov2016instance} to normalize convolutional feature maps.
\newline

\noindent \textbf{Learning and inference.}
We use the Adam solver \cite{kingma2014adam} with a learning rate of $0.0002$ and coefficients of (0.5, 0.999), where the latter is used to compute running averages of gradients and their squares. The learning rate is fixed in the initial 100 epochs, and linearly decays to zero over the next 100 epochs. Following CycleGAN, the negative log likelihood objective is replaced with a more stable and effective least-squares loss \cite{mao2017least} for $\mathcal{L}_{gan}$. The discriminator is updated with random samples from a history of generated images stored in an image buffer \cite{shrivastava2017learning} of size 50. The generator and discriminator are optimized alternately. In the inference phase, we feed an image only into the learned generator $G_{XY}$ to obtain a translated image. 

\setlength\tabcolsep{4.4pt}
\begin{table*}[t]
\centering
\begin{tabular}{ c || c  c  c | c  c  c }
\hline
\multirow{ 2 }{*}{method} & \multicolumn{3}{   c | }{ image $\to$ parsing } & \multicolumn{3}{   c  }{ parsing $\to$ image} \\ \cline{2-7}
 & pixel acc & class acc & mean IoU  & pixel acc & class acc & mean IoU \\
\hline\hline
\rowcolor{myazure}
\multicolumn{7}{   c  }{Benchmark Performance} \\
\hline
CoGAN \cite{liu2016coupled} & 0.45 & 0.11 & 0.08 & 0.40 & 0.10 & 0.06 \\
BiGAN/ALI \cite{donahue2016adversarial, dumoulin2016adversarially} & 0.41 & 0.13 & 0.07 & 0.19 & 0.06 & 0.02 \\
SimGAN \cite{shrivastava2017learning} & 0.47 & 0.11 & 0.07 & 0.20 & 0.10 & 0.04 \\
CycleGAN (Cycle) \cite{zhu2017unpaired} & \bf 0.58 & 0.22 & 0.16 & 0.52 & 0.17 & 0.11 \\
DistanceGAN \cite{benaim2017one} & - & - & - & 0.53 & 0.19 & 0.11 \\
\rowcolor{mygray}
GAN alone (baseline) & 0.514 & 0.160 & 0.104 & 0.437 & 0.161 & 0.098 \\
\hline
GcGAN-\emph{rot} & 0.574 & \bf 0.234 & 0.170 & \bf 0.551 & \bf 0.197 & \bf 0.129 \\
GcGAN-\emph{vf} & 0.576 & 0.232 & \bf 0.171 & 0.548 & 0.196 & 0.127 \\

\hline\hline
\rowcolor{myhoneydew}
\multicolumn{7}{   c  }{Ablation Studies (\textcolor{blue}{Robustness \& Compatibility})} \\
\hline
GcGAN-\emph{rot}-Seperate & 0.575 & 0.233 & 0.170 & 0.545 & 0.196 & 0.124 \\
GcGAN-Mix & 0.573 & 0.229 & 0.168 & 0.545 & 0.197 & 0.128 \\
GcGAN-\emph{rot} + Cycle & \bf 0.587 & \bf 0.246 & \bf 0.182 & \bf 0.557 & \bf 0.201 & \bf 0.132 \\
\hline
\end{tabular}
\captionsetup{font={small}}
\caption{\textbf{Parsing scores on Cityscapes.} GcGAN-emph{rot}-Separate: $G_{XY}$ and $G_{\tilde{X}\tilde{Y}}$ do not share parameters. GcGAN-Mix: GcGAN with a mixture of transformations (\emph{rot} and \emph{vf}). GcGAN-\emph{rot} + Cycle: GcGAN-\emph{rot} with the cycle-consistency constraint.}
\label{tab:cityscapes}
\vspace{-0.3cm}
\end{table*}

\section{Experiments}
We apply our GcGAN to a wide range of applications and make both quantitative and qualitative comparisons with the baseline (GAN alone) and previous state-of-the-art methods including DistanceGAN and CycleGAN. We also study different ablations (based on \emph{rot}) to analyze our geometry-consistency constraint. Since adversarial networks are not always stable, every independent experiment could result in slightly different scores. The scores in the quantitative analysis are computed by the average  on three independent experiments. 

\subsection{Quantitative Analysis}
The results demonstrate that our geometry-consistency constraint can not only partially filter out the candidate solutions having mode collapse or semantic distortions and thus produce more sensible translations, but also compatible with other unsupervised constraints such as cycle consistency \cite{zhu2017unpaired} and distance preservation \cite{benaim2017one}.
\newline

\noindent \textbf{Cityscapes.} Cityscapes \cite{Cordts2016Cityscapes} contains 3975 image-label pairs, with 2975 used for training and 500 for validation (test in this paper). For a fair comparison with CycleGAN, the translators are trained at a resolution of $128 \times 128$ in an unaligned fashion. We evaluate our domain mappers using FCN scores and scene parsing metrics following previous works \cite{long2015fully, Cordts2016Cityscapes, zhu2017unpaired}. Specifically, for parsing $\to$ image, we assume that a high-quality translated image should produce qualitative semantic segmentation like real images when feeding it into a scene parser. Thus, we employ the pretrained FCN-8s \cite{long2015fully} provided by pix2pix \cite{isola2017image} to predict semantic labels for the 500 translated images. The label maps are then resized to the original resolution of $1024 \times 2048$ and compared against the ground truth labels using some standard scene parsing metrics including pixel accuracy, class accuracy, and mean IoU \cite{long2015fully}. For image $\to$ parsing, since the fake labels are in the RGB format, we simply convert them into class-level labels using the nearest neighbor search strategy. In particular, we have 19 (category labels) $+$ 1 (ignored label) categories for Cityscapes, each with a corresponding color value (RGB). For a pixel $i$ in a translated parsing, we compute the distances between the 20 groundtruth color values and the color value of pixel $i$. The label of pixel $i$ should be the one with the smallest distance. Then, the aforementioned metrics are used to evaluate our mapping on the 19 category labels. 

The parsing scores for both image $\to$ parsing and parsing $\to$ image tasks are presented in Table~\ref{tab:cityscapes}. Our GcGAN outperforms the baseline (GAN alone) by a large margin.
We take the average of pixel accuracy, class accuracy, and mean IoU as the final score for analysis \cite{zhou2017scene}, {\it i.e}., $\text{score} = (\text{pixel acc} + \text{class acc} + \text{mean IoU})/3$.  For image $\to$ parsing, GcGAN ($32.6\%$) yields a slightly higher score than CycleGAN ($32.0\%$). For parsing $\to$ image, GcGAN ($29.0\%\sim29.5\%$) obtains a convincing improvement of $1.3\%\sim1.8\%$ over the state-of-the-art approach distanceGAN ($27.7\%$). 

We next perform ablation studies to investigate the robustness and compatibility of GcGAN, including GcGAN-\emph{rot}-Seperate, GcGAN-Mix, and GcGAN-\emph{rot} + Cycle. The scores are reported in Table~\ref{tab:cityscapes}. Specifically, GcGAN-\emph{rot}-Seperate shows that the generator $G_{XY}$ employed in GcGAN is sufficient to handle both the style transfers (without shape deformation) $X \to Y$ and $\tilde{X} \to \tilde{Y}$. GcGAN-Mix demonstrates that persevering a geometric transformation can filter out most of the candidate solutions having mode collapse or undesired shape deformation, but preserving more ones can not leach more. For GcGAN-\emph{rot} + Cycle, we set the trade-off parameter for $\mathcal{L}_{cyc}$ to $10.0$ as published in CycleGAN. The consistent improvement is a credible support that our geometry-consistency constraint is compatible with the widely-used cycle-consistency constraint.

\begin{table*}[h]
\centering
\begin{tabular}{ c | c   }
\hline
method  & class acc ($\%$) \\
 \hline\hline
 \rowcolor{myazure}
 \multicolumn{2}{   c  }{  Benchmark Performance} \\ 
 \hline
 DistanceGAN (Dist.) \cite{benaim2017one} & 26.8 \\
 CycleGAN (Cycle) \cite{zhu2017unpaired} & 26.1  \\
 Self-Distance \cite{benaim2017one}  & 25.2  \\
 \hline
GcGAN-\emph{rot}  & 32.5  \\
GcGAN-\emph{vf}  & \bf{33.3} \\
\hline\hline
\rowcolor{myhoneydew}
\multicolumn{2}{  c  }{  Ablation Studies (\textcolor{blue}{Compatibility}) } \\
\hline
\rowcolor{mygray}
Cycle + Dist. \cite{benaim2017one} & 18.0 \\
GcGAN-\emph{rot} + Dist. & \bf{34.0} \\
GcGAN-\emph{rot} + Cycle & 33.8 \\
GcGAN-\emph{rot} + Dist. + Cycle & 33.2 \\
\hline
\end{tabular}

\captionsetup{font={small}}
\caption{Quantitative scores for SVHN $\to$ MNIST.}
\label{tab:score-mnist}
\end{table*}

\noindent \textbf{SVHN $\to$ MNIST}. We then apply our approach to the SVHN $\to$ MNIST translation task. The translation models are trained on 73257 and 60000 training images of resolution $32 \times 32$  contained in the SVHN and MNIST training sets, respectively. The experimental settings follow DistanceGAN \cite{benaim2017one}, including the default trade-off parameters for $\mathcal{L}_{cyc}$ and $\mathcal{L}_{dis}$, and the network architectures  for the generators and the discriminators. We compare our GcGAN with both DistanceGAN and CycleGAN in this translation task. To obtain quantitative results, we feed the translated images into a pretrained classifier trained on the MNIST training split, as done in \cite{benaim2017one}. Note that, the experimental settings for domain mapping (GcGAN) and domain adaptation are totally different, so is the captured classification accuracy. Domain adaptation methods have access to the source domain digit labels while image translation does not.

Classification accuracies are reported in Table~\ref{tab:score-mnist}. Both GcGAN-\emph{rot} and GcGAN-\emph{vf} outperform DistanceGAN and CycleGAN by a large margin (about $6\% \sim 7\%$). From the ablations, adding our geometry-consistency constraint to current unsupervised domain mapping frameworks will achieve different levels of improvements against the original ones. Note that, it seems that the distance-preservation constraint is not compatible with the cycle-consistency constraint, but our geometry-consistency constraint can improve both ones.
\begin{figure*}[h]
\begin{center}
\begin{subfigure}{0.49\textwidth}
  \begin{center}
  \scalebox{1}[1]{\includegraphics[height=3cm, width=6cm]{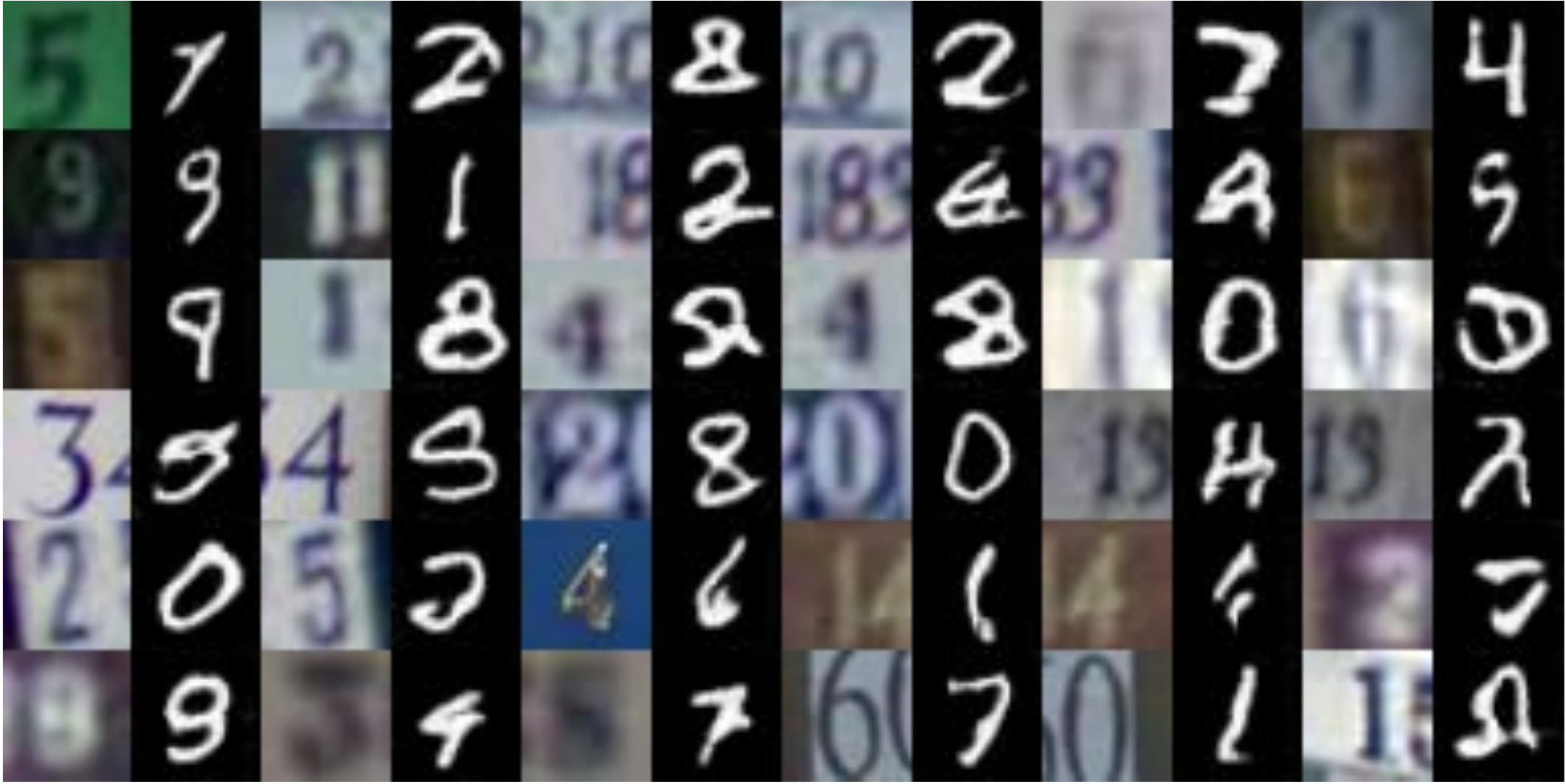}}
  \caption*{DistanceGAN \cite{benaim2017one}}
  \end{center}
\end{subfigure}%
\begin{subfigure}{0.49\textwidth}
  \begin{center}
  \scalebox{1}[1]{\includegraphics[height=3cm, width=6cm]{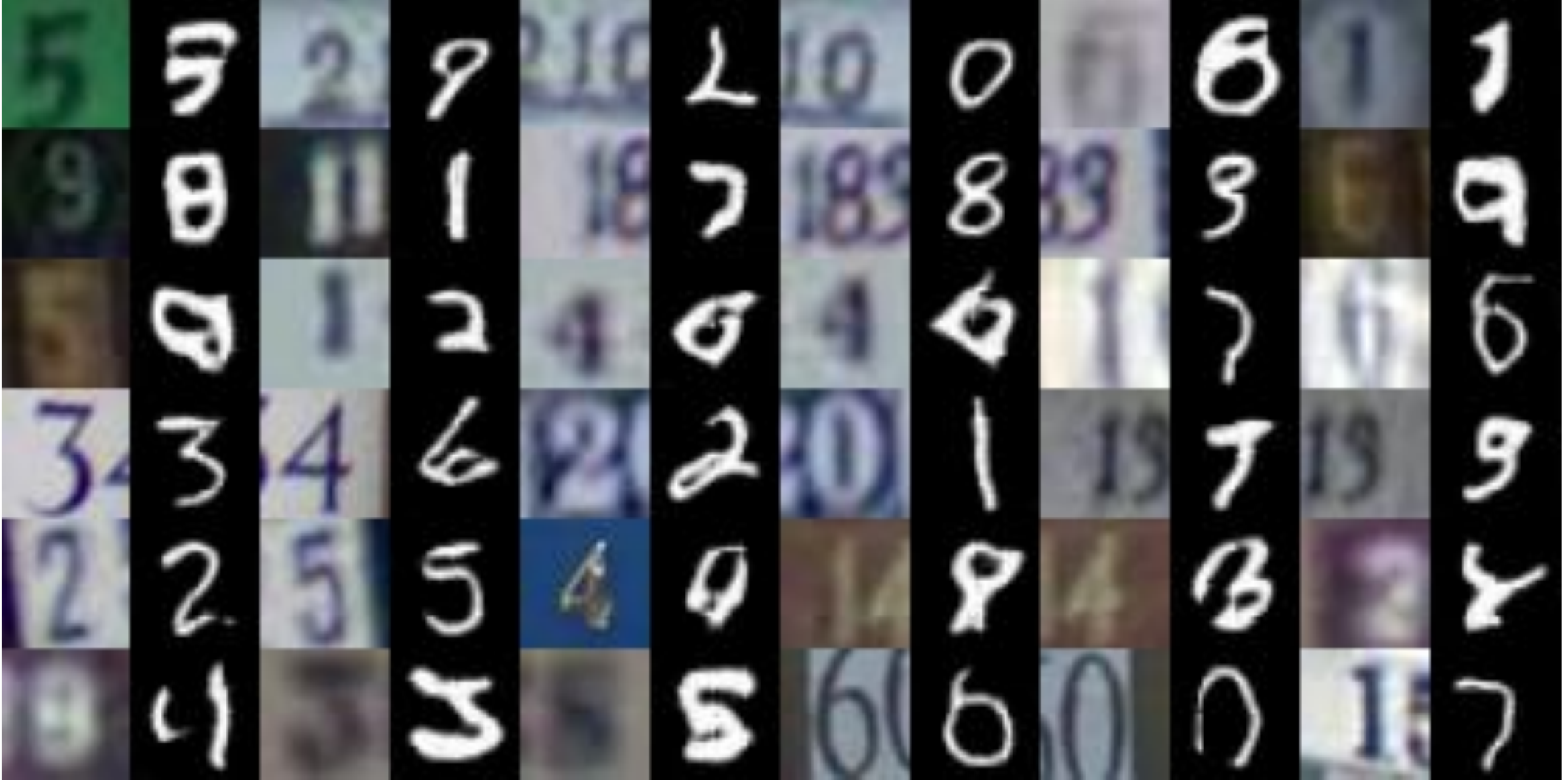}}
  \caption*{GcGAN}
  \end{center}
\end{subfigure}%
\captionsetup{font={small}}
\caption{Qualitative comparison for SVHN $\to$ MNIST.}
\label{fig:mnist}
\end{center}
\vspace{-0.4cm}
\end{figure*}

\noindent \textbf{Google Maps}. We obtain 2194 $(\text{map}, \text{aerial photo})$ pairs of images in and around New York City from Google Maps \cite{isola2017image}, and split them into training and test sets with 1096 and 1098 pairs, respectively. We train Map $\rightleftharpoons$ Aerial photo translators with an image size of $256 \times 256$ using the training set in an unsupervised manner (unpaired) by ignoring the pair information.  For Aerial photo $\to$ Map, we make comparisons with CycleGAN using average RMSE and pixel accuracy ($\%$). Given a pixel $i$ with the ground-truth RGB value $(r_i, g_i, b_i)$ and the predicted RGB value $(r'_i, g'_i, b'_i)$, if $max(|r_i - r'_i|, |g_i - g'_i|, |b_i - b'_i|) < \delta$, we consider this is an accurate prediction. Since maps only contain a limited number of different RGB values, it is reasonable to compute pixel accuracy using this strategy ($\delta_1 = 5$ and $\delta_2 = 10$ in this paper). For Map $\to$ Aerial photo, we only show some qualitative results in Figure~\ref{fig:cityscape}. 

\begin{table*}[h]
\centering
\begin{tabular}{ c || c | c  c  }
\hline
method  & RMSE & acc $(\delta_1)$ &  acc $(\delta_2)$ \\
 \hline\hline
 \rowcolor{myazure}
 \multicolumn{4}{   c  }{  Benchmark Performance} \\ 
 \hline
 CycleGAN \cite{zhu2017unpaired} &  \bf 28.15 & \bf 41.8 & \bf 63.7 \\
 \rowcolor{mygray}
 GAN alone (baseline)  &  33.27 & 19.3 & 42.0 \\
 \hline
GcGAN-\emph{rot}   & 28.31 & 41.2 & 63.1 \\
GcGAN-\emph{vf}  &  28.50 & 37.3 & 58.9 \\
\hline\hline
\rowcolor{myhoneydew}
\multicolumn{4}{  c  }{  Ablation Studies (\textcolor{blue}{Robustness \& Compatibility}) } \\
\hline
GcGAN-\emph{rot}-Separate  &  30.25 & 40.7 & 60.8 \\
GcGAN-Mix  &  \bf 27.98 & \bf 42.8 & \bf 64.6 \\
GcGAN-\emph{rot} + Cycle  & 28.21 & 40.6 & 63.5 \\
\hline
\end{tabular}
\captionsetup{font={small}}
\caption{Quantitative scores for Aerial photo $\to$ Map.}
\label{tab:score-map}
\end{table*}

From the scores presented in Table~\ref{tab:score-map}, it can be seen that GcGAN produces superior translations to the baseline (GAN alone). In particular, GcGAN yields an $18.0\% \sim 21.9\%$ improvement over the baseline with respect to pixel accuracy when $\delta = 5.0$, demonstrating that the fake maps obtained by our GcGAN contain more details. In addition, our one-sided GcGANs achieve competitive even slightly better scores compared with the two-sided CycleGAN.

\begin{figure*}[ht!]
\begin{center}
\begin{subfigure}{0.99\textwidth}
  \begin{center}
  \scalebox{1}[1]{\includegraphics[scale=0.8]{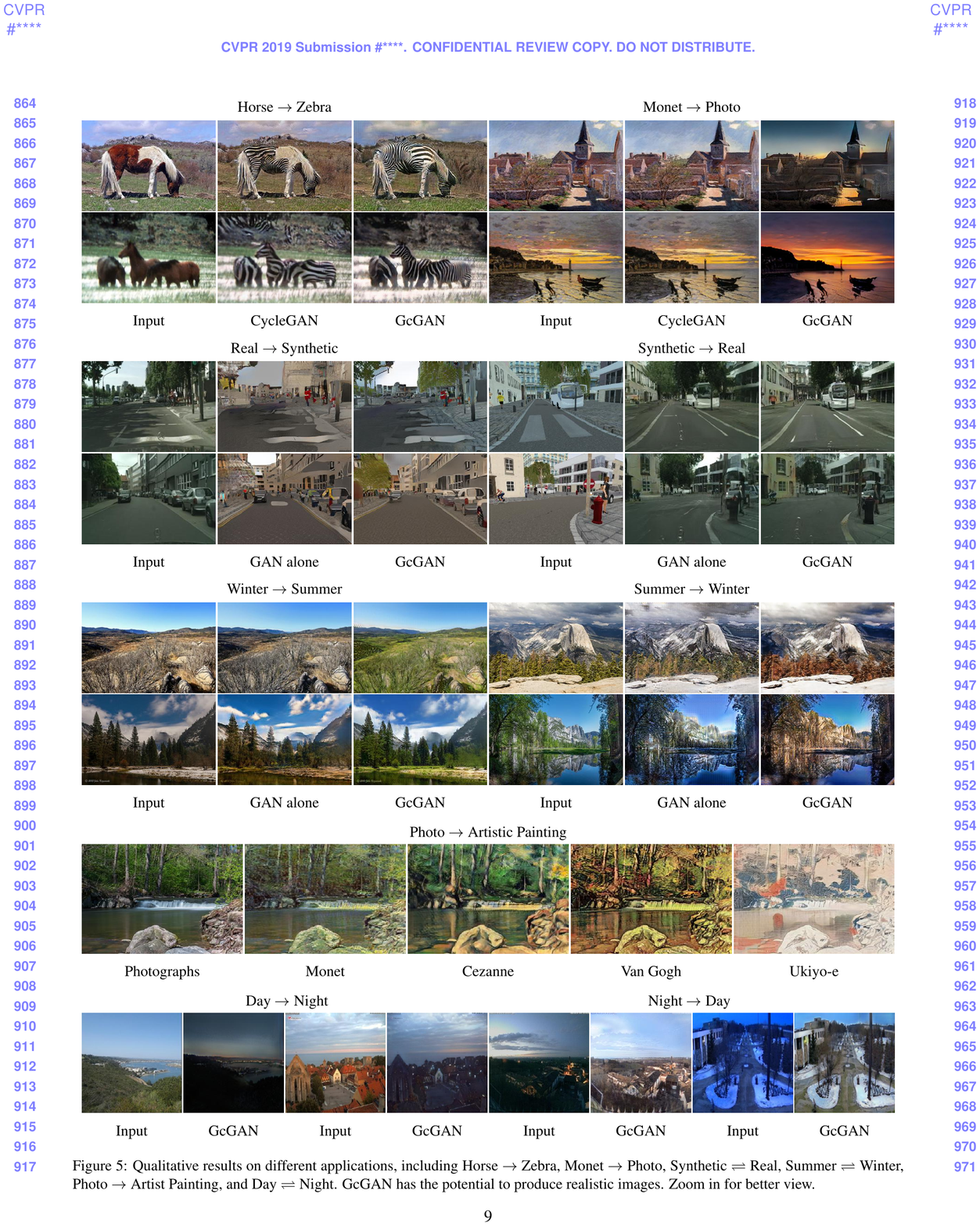}}
  \end{center}
\end{subfigure}%

\captionsetup{font={small}}
\caption{Qualitative results on different applications, including Horse $\to$ Zebra, Monet $\to$ Photo, Synthetic $\rightleftharpoons$ Real, Summar $\rightleftharpoons$ Winter, Photo $\to$ Artist Painting, and Day $\rightleftharpoons$ Night. GcGAN has the potential to produce realistic images. Zoom in for better view.}
\label{fig:applications}
\end{center}
\vspace{-0.5cm}
\end{figure*}

\subsection{Qualitative Evalutation}
The qualitative results are shown in Figure~\ref{fig:cityscape}, Figure~\ref{fig:mnist}, and Figure \ref{fig:applications}. While GAN alone suffers from mode collapse, our geometry-consistency constraint can provide an effective remedy, thus helps to generate empirically more impressive translations on various applications. The following applications are trained in the image size of $256 \times 256$ with the \emph{rot} geometric transformation.

\noindent \textbf{Horse $\to$ Zebra}. We apply GcGAN to the widely studied object transfiguration application task, {\it i.e}., Horse $\to$ Zebra. The images are randomly sampled from ImageNet \cite{deng2009imagenet} using the keywords ({\it i.e}., wild horse and zebra). The numbers of training images are 939 and 1177 for horse and zebra, respectively. We find that training GcGAN without parameter sharing would produce preferable translations for the task.

\noindent \textbf{Synthetic $\rightleftharpoons$ Real}. We employ the 2975 training images from Cityscapes as the real-world scenes, and randomly select 3060 images from SYNTHIA-CVPR16 \cite{Ros_2016_CVPR}, which is a virtual urban scene benchmark, as the synthetic images. 

\noindent \textbf{Summer $\rightleftharpoons$ Winter}. The images used for the season translation tasks are provided by CycleGAN. The training set sizes for Summer and Winter are 1273 and 854. 

\noindent \textbf{Photo $\rightleftharpoons$ Artistic Painting}. We translate natural images to artistic paintings with different art styles, including Monet, Cezanne, Van Gogh, and Ukiyo-e. We also perform GcGAN on the translation task of Monet's paintings $\to$ photographs. We use the photos and paintings (Monet: 1074, Cezanne: 584, Van Gogh: 401, Ukiyo-e: 1433, and Photographs: 6853) collected by CycleGAN for training. 

\noindent \textbf{Day $\rightleftharpoons$ Night.} We randomly extract 4500 training images for both Day and Night from the 91 webcam sequences captured by \cite{laffont2014transient}. 


\section{Conclusion}
In this paper, we propose to enforce geometry consistency as a constraint, which can be viewed as a predefined geometric transformation $f(\cdot)$ preserving the geometry of a scene for unsupervised domain mapping. The geometry-consistency constraint makes the translation networks on the original images and transformed images co-regularize each other, which not only provides an effective remedy to the mode collapse problem suffered by standard GANs, but also reduces the semantic distortions in the translation. We evaluate our model, {\it i.e.}, the geometry-consistent generative adversarial network (GcGAN), both qualitatively and quantitatively in various applications. Our experimental results demonstrate that GcGAN achieves competitive and sometimes even better translations than the state-of-the-art methods including DistanceGAN and CycleGAN. Finally, our geometry-consistency constraint is compatible with other well-studied unsupervised constraints. 

{\small
\bibliographystyle{ieee}
\bibliography{egbib_cvpr}
}


\newpage
\section*{Network Architecture}
The generator and discriminator (except for SVHN $\to$ MNIST) presented before are shown in Tab.~\ref{tab:network}. For convenience, we use the following abbreviation: C = Feature channel, K = Kernel size, S = Stride size, Deconv/Conv = Deconvolutional/Convolutional layer, and ResBlk = A residual block.

\begin{table}[h]
\caption{The generator and discriminator used in our experiments ($256 \times 256$).}
\centering
\begin{tabular}{ c | c | c   c   c   }
\hline
\multicolumn{5}{   c  }{ Generator } \\
\hline
\hline
Index & Layer & C & K & S  \\
\hline
1 & Conv + ReLU & 64 & 7 & 1 \\
2 & Conv  + ReLU & 128 & 3 & 2 \\
3 & Conv + ReLU & 256 & 3 & 2 \\
4 & ResBlk + ReLU & 256 & 3 & 1 \\
5 & ResBlk + ReLU & 256 & 3 & 1 \\
6 & ResBlk + ReLU & 256 & 3 & 1 \\
7 & ResBlk + ReLU & 256 & 3 & 1 \\
8 & ResBlk + ReLU & 256 & 3 & 1 \\
9 & ResBlk + ReLU & 256 & 3 & 1 \\
10 & ResBlk + ReLU & 256 & 3 & 1 \\
11 & ResBlk + ReLU & 256 & 3 & 1 \\
12 & ResBlk + ReLU & 256 & 3 & 1 \\
12 & Deconv + ReLU & 128 & 3 & 2 \\
13 & Deconv + ReLU & 64 & 3 & 2 \\
14 & Conv & 3 & 7 & 1 \\
15 & Tanh & - & - & - \\
\hline
\multicolumn{5}{   c  }{ Discriminator } \\
\hline
\hline
1 & Conv + LeakyReLU & 64 & 4 & 2 \\
2 & Conv + LeakyReLU & 128 & 4 & 2 \\
3 & Conv + LeakyReLU & 256 & 4 & 2 \\
4 & Conv + LeakyReLU & 512 & 4 & 1\\
5 & Conv  & 512 & 4 & 1 \\
\hline
\end{tabular}
\label{tab:network}
\end{table}

The network architecture for SVHN $\to$ MNIST is reported in Tab.~\ref{tab:net-mnist}.

\begin{table}[h]
\caption{The network architecture for SVHN $\to$ MNIST.}
\centering
\begin{tabular}{ c | c | c   c   c   }
\hline
\multicolumn{5}{   c  }{ Generator } \\
\hline
\hline
Index & Layer & C & K & S  \\
\hline
1 & Conv + LeakyReLU & 64 & 4 & 2 \\
2 & Conv  + LeakyReLU & 128 & 4 & 2 \\
3 & Conv + LeakyReLU & 128 & 3 & 1 \\
4 & Conv + LeakyReLU & 128 & 3 & 1 \\
5 & Deconv + LeakyReLU & 64 & 4 & 2 \\
5 & Deconv + LeakyReLU & 1 & 4 & 2 \\
15 & Tanh & - & - & - \\
\hline
\multicolumn{5}{   c  }{ Discriminator } \\
\hline
\hline
1 & Conv + LeakyReLU & 64 & 4 & 2 \\
2 & Conv + LeakyReLU & 128 & 4 & 2 \\
3 & Conv + LeakyReLU & 256 & 4 & 2 \\
4 & Conv + LeakyReLU & 512 & 4 & 1\\
5 & Conv  & 512 & 4 & 1 \\
\hline
\end{tabular}
\label{tab:net-mnist}
\end{table}

\newpage
\begin{figure*}[t]
\begin{center}
\begin{subfigure}{0.98\textwidth}
  \begin{center}
  \scalebox{1}[1]{\includegraphics[scale=1.0]{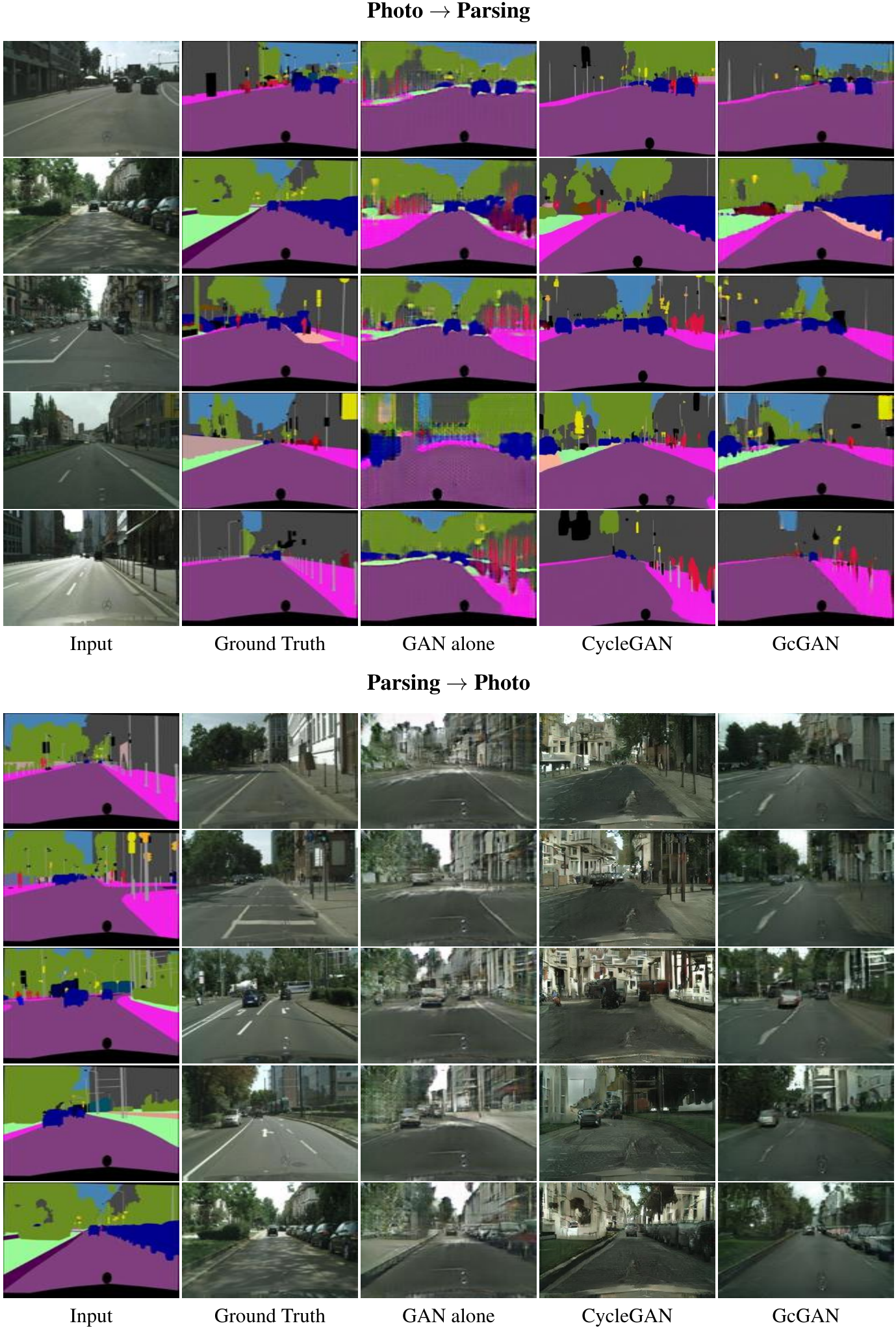}}
  \end{center}
\end{subfigure}%
\captionsetup{font={small}}
\caption{\textbf{Cityscapes (Parsing $\rightleftharpoons$ Image).} The results for CycleGAN [\textcolor{green}{62}] are produced by the officially provided pretrained PyTorch models. GcGAN denotes GcGAN-\emph{rot}.}
\label{fig:supp_1}
\end{center}
\end{figure*}

\begin{figure*}[t]
\begin{center}
\begin{subfigure}{0.98\textwidth}
  \begin{center}
  \scalebox{1}[1]{\includegraphics[scale=1.0]{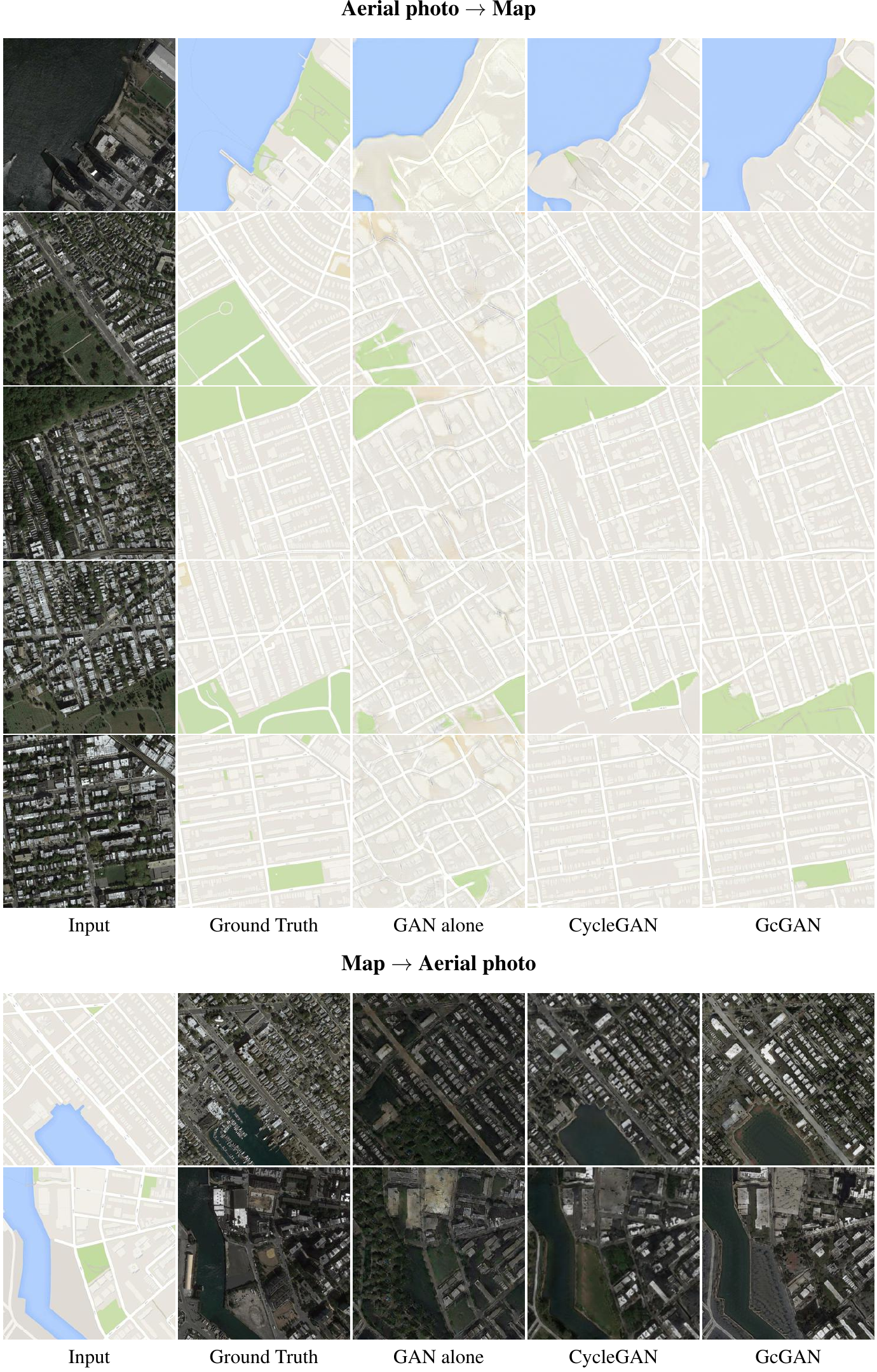}}
  \end{center}
\end{subfigure}%
\captionsetup{font={small}}
\caption{\textbf{Google Mpas (Aerial photo $\rightleftharpoons$ Map).} For Map $\to$ Aerial photo,  GcGAN produces competitive translations compared with CycleGAN [\textcolor{green}{62}].}
\label{fig:supp_2}
\end{center}
\end{figure*}

\begin{figure*}[t]
\begin{center}
\begin{subfigure}{0.98\textwidth}
  \begin{center}
  \scalebox{1}[1]{\includegraphics[scale=1.0]{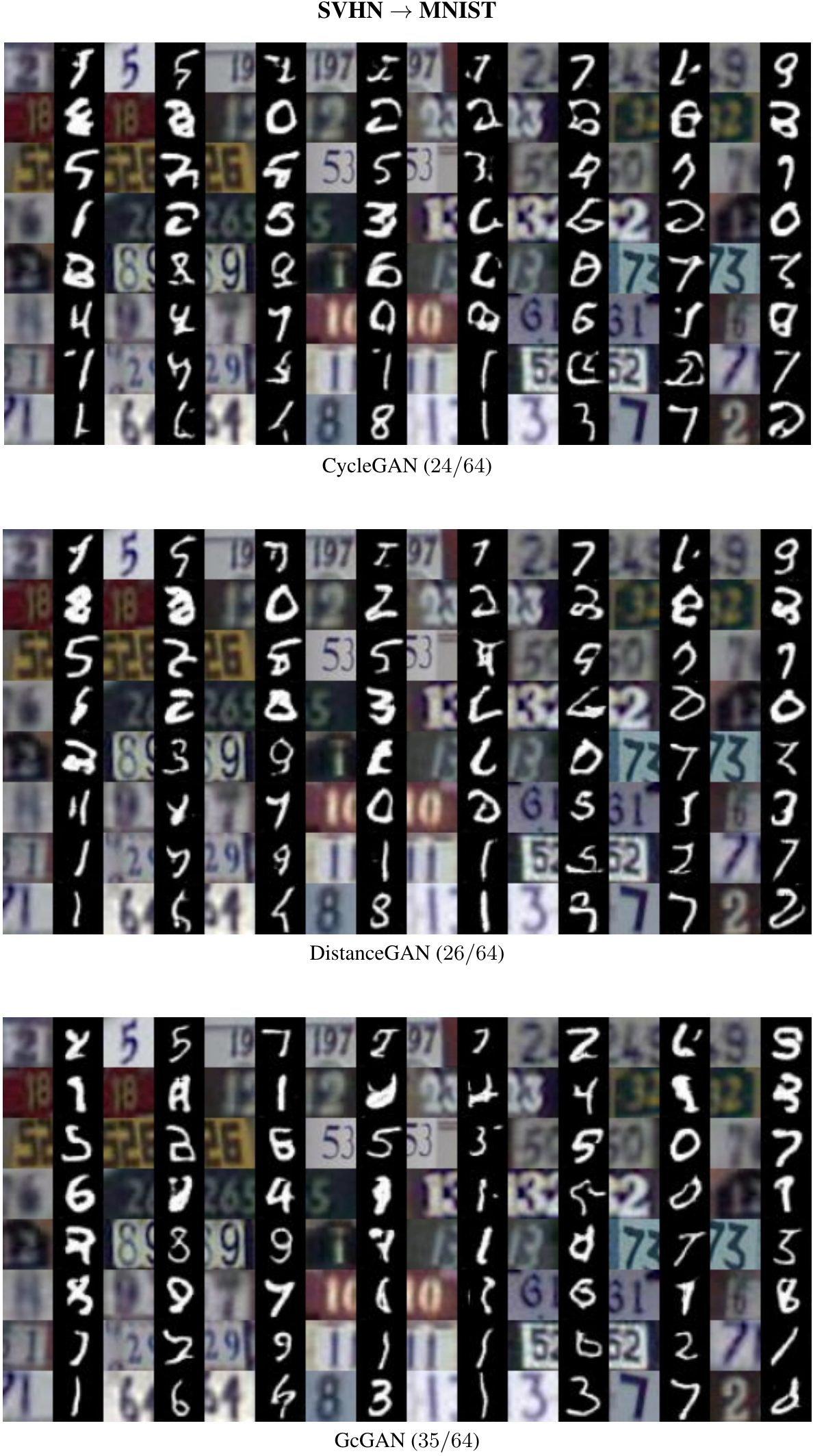}}
  \end{center}
\end{subfigure}%
\captionsetup{font={small}}
\caption{\textbf{SVHN $\to$ MNIST.} The qualitative results for both CycleGAN [\textcolor{green}{62}] and DistanceGAN [\textcolor{green}{5}] come from DistanceGAN [\textcolor{green}{5}]. The correct translations are about 24, 26, and 35 for CycleGAN, DistanceGAN, and GcGAN, respectively.}
\label{fig:supp_3}
\end{center}
\end{figure*}

\begin{figure*}[t]
\begin{center}
\begin{subfigure}{0.98\textwidth}
  \begin{center}
  \scalebox{1}[1]{\includegraphics[scale=1.0]{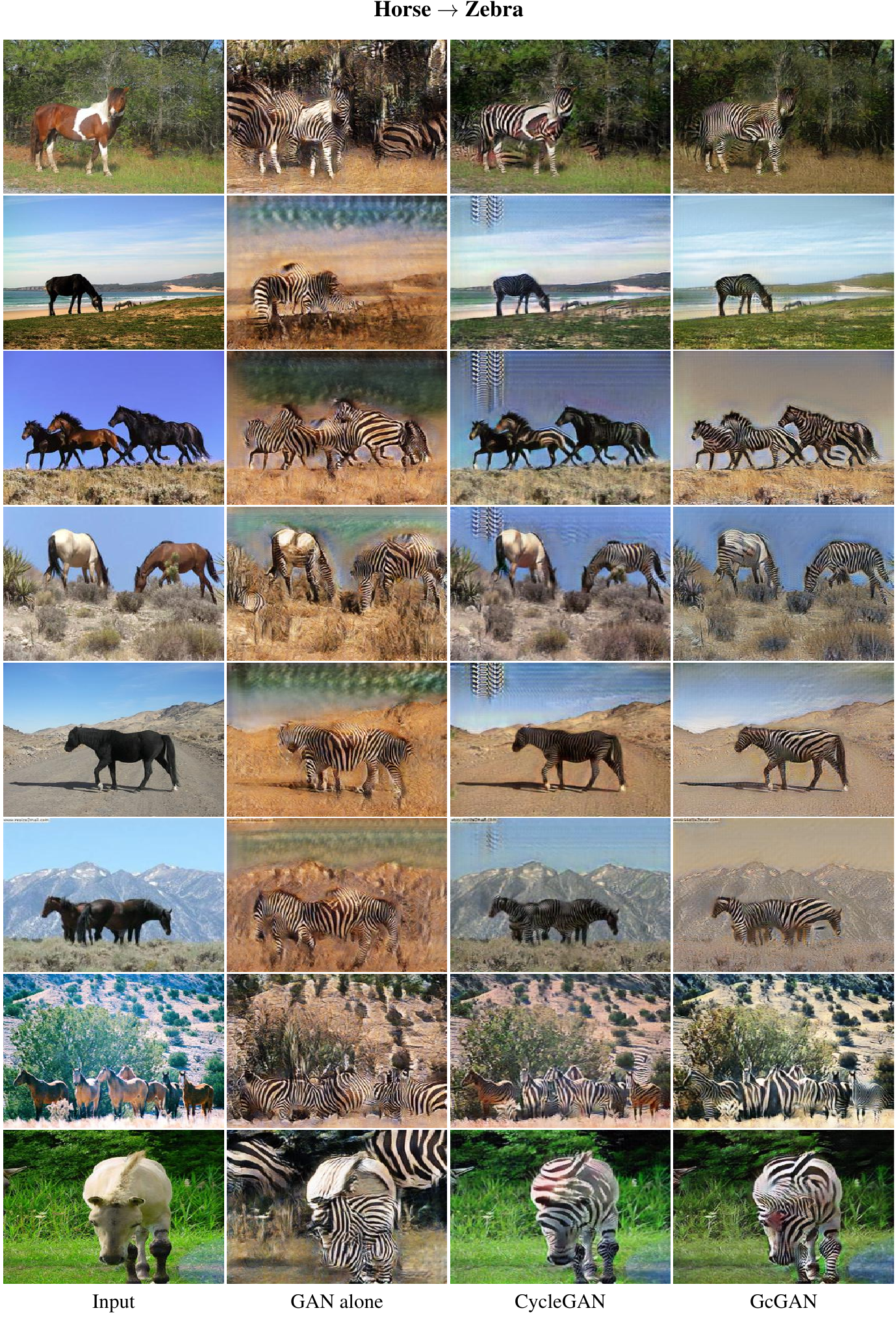}}
  \end{center}
\end{subfigure}%
\captionsetup{font={small}}
\caption{\textbf{Horse $\to$ Zebra.} For this task, GcGAN generates slightly better translations for some images, but can not perform better than CycleGAN [\textcolor{green}{62}] generally.}
\label{fig:supp_4}
\end{center}
\end{figure*}

\begin{figure*}[t]
\begin{center}
\begin{subfigure}{0.98\textwidth}
  \begin{center}
  \scalebox{1}[1]{\includegraphics[scale=1.0]{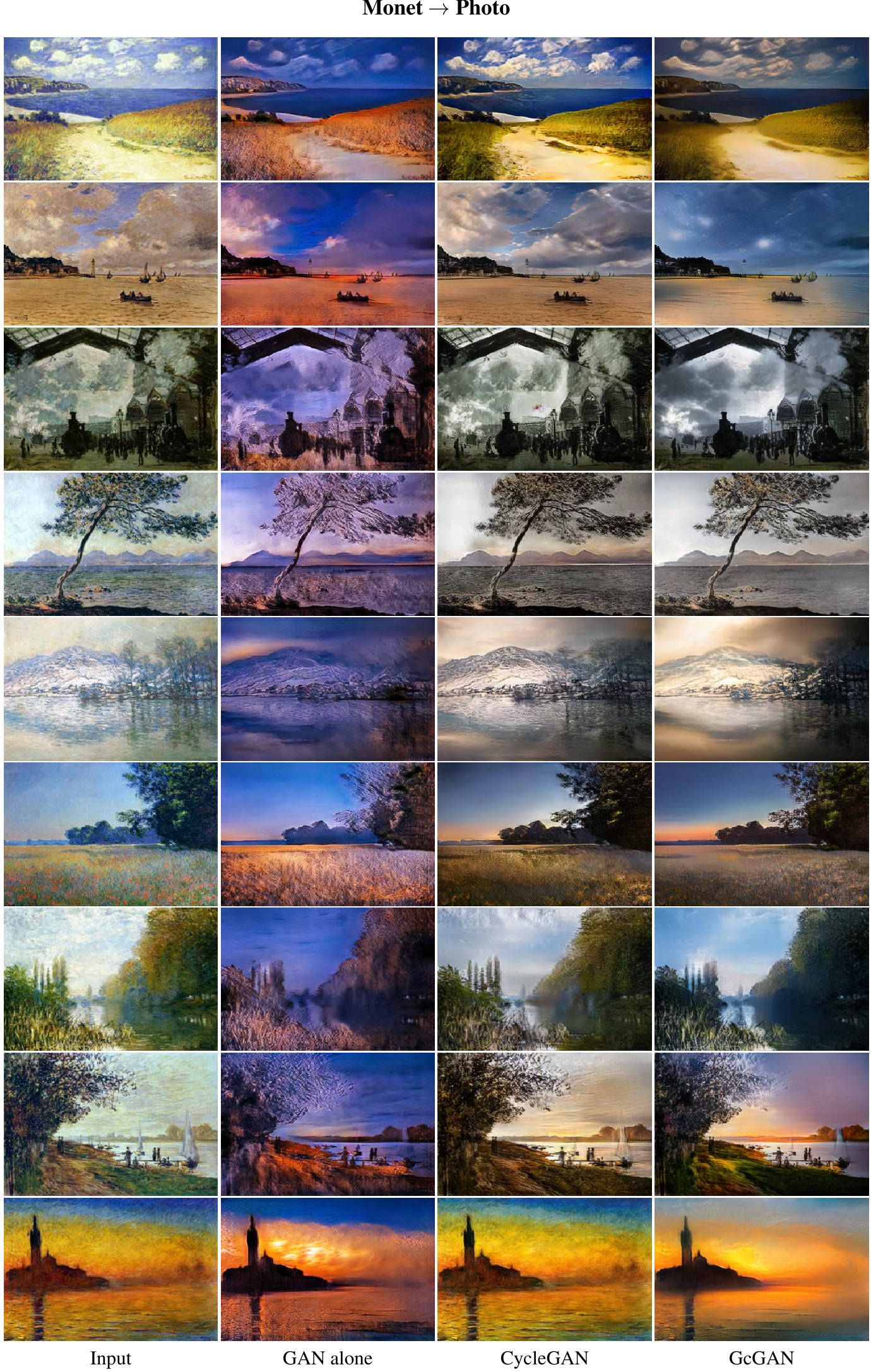}}
  \end{center}
\end{subfigure}%
\captionsetup{font={small}}
\caption{\textbf{Monet $\to$ Photo.} GcGAN is superior in generating realistic images. Zoom in for better view.}
\label{fig:supp_5}
\end{center}
\end{figure*}

\begin{figure*}[t]
\begin{center}
\begin{subfigure}{0.98\textwidth}
  \begin{center}
  \scalebox{1}[1]{\includegraphics[scale=1.0]{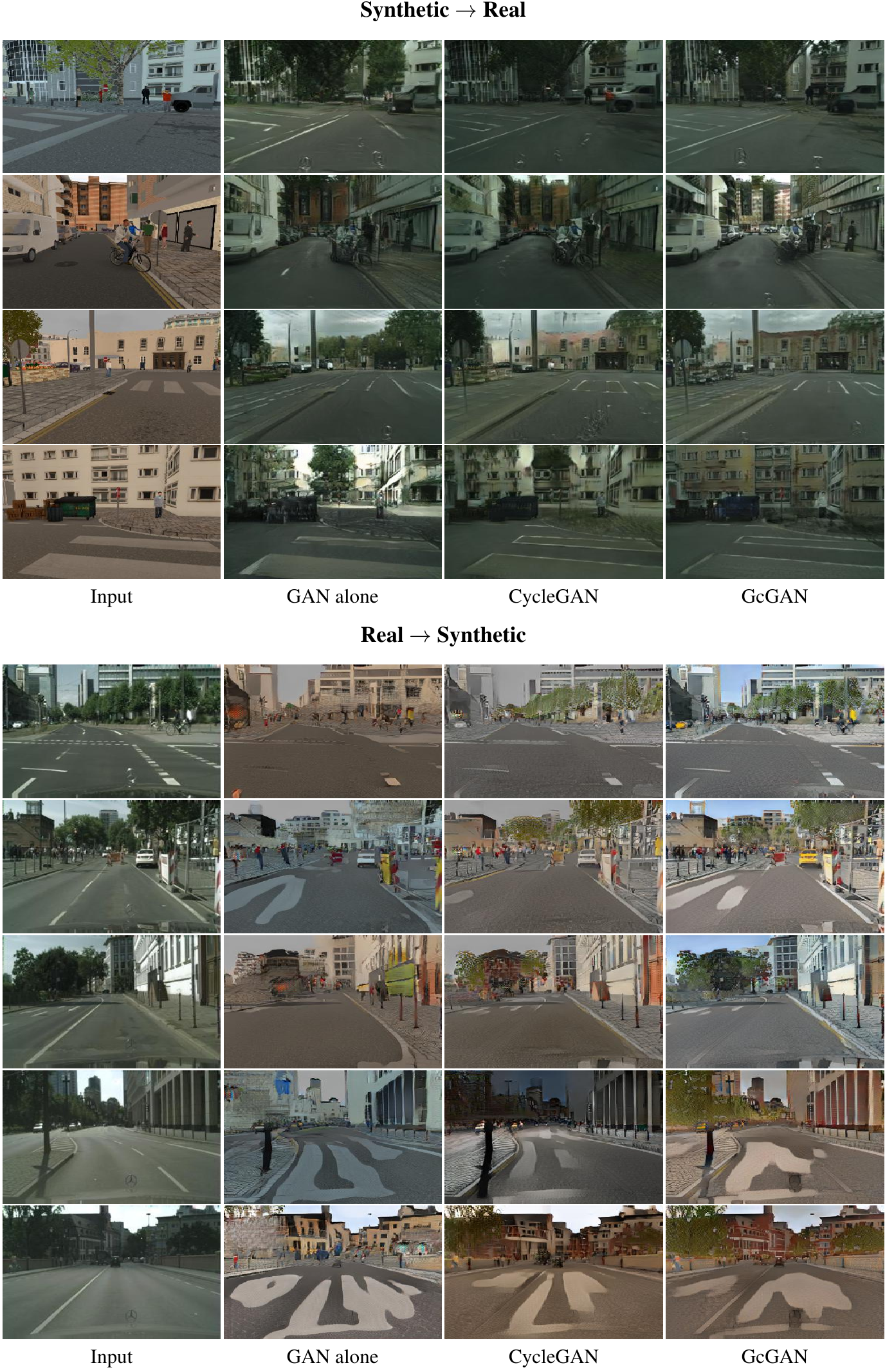}}
  \end{center}
\end{subfigure}%
\captionsetup{font={small}}
\caption{\textbf{Synthetic $\rightleftharpoons$ Real.} We train CycleGAN [\textcolor{green}{62}] using the released PyTorch codes. The results produced by GcGAN contain more details. Zoom in for better view.}
\label{fig:supp_7}
\end{center}
\end{figure*}

\begin{figure*}[t]
\begin{center}
\begin{subfigure}{0.98\textwidth}
  \begin{center}
  \scalebox{1}[1]{\includegraphics[scale=1.0]{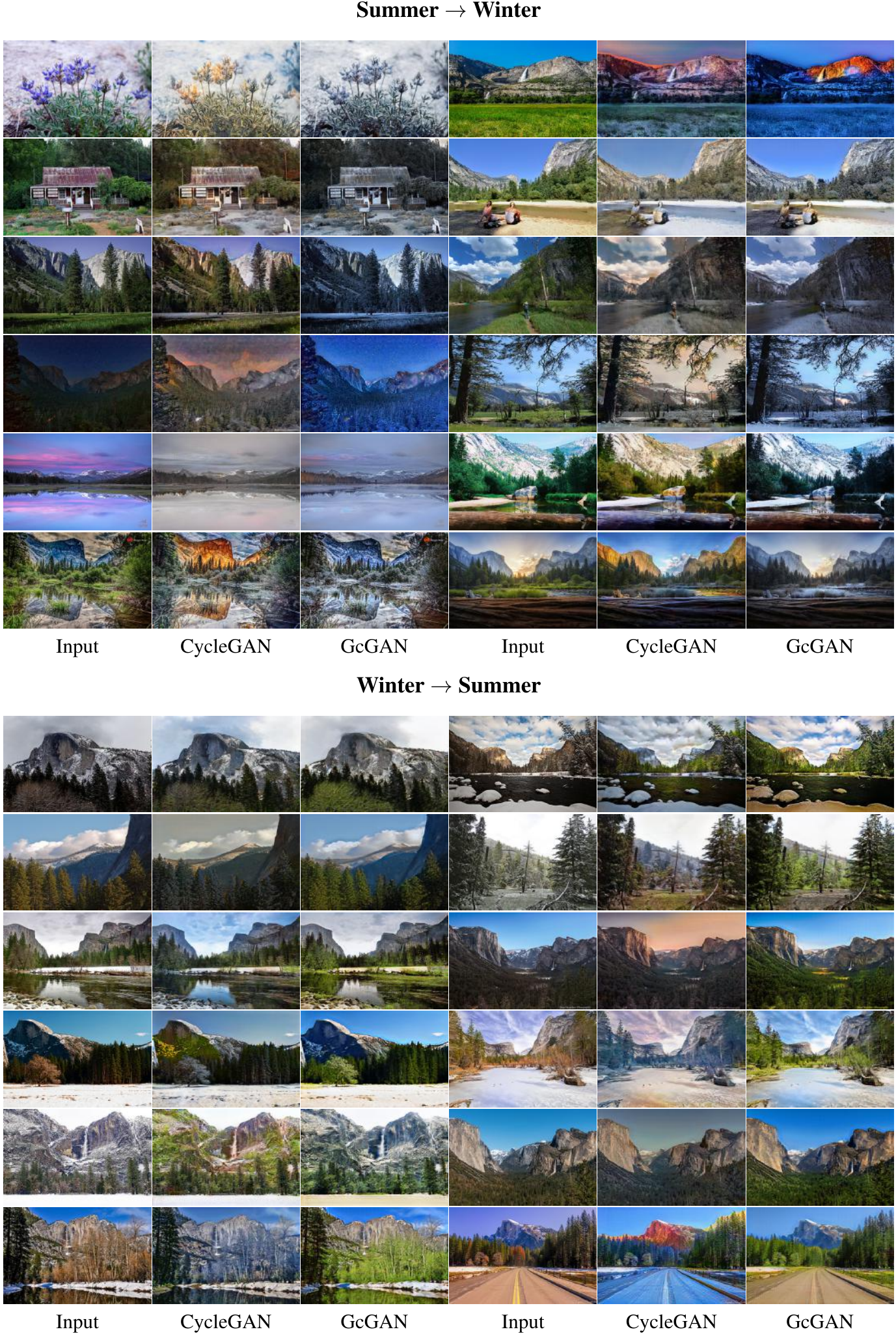}}
  \end{center}
\end{subfigure}%
\captionsetup{font={small}}
\caption{\textbf{Summer $\rightleftharpoons$ Winter.} Here, GcGAN represents GcGAN-\emph{rot}-Seperate. Zoom in for better view.}
\label{fig:supp_9}
\end{center}
\end{figure*}

\begin{figure*}[t]
\begin{center}
\begin{subfigure}{0.98\textwidth}
  \begin{center}
  \scalebox{1}[1]{\includegraphics[scale=1.0]{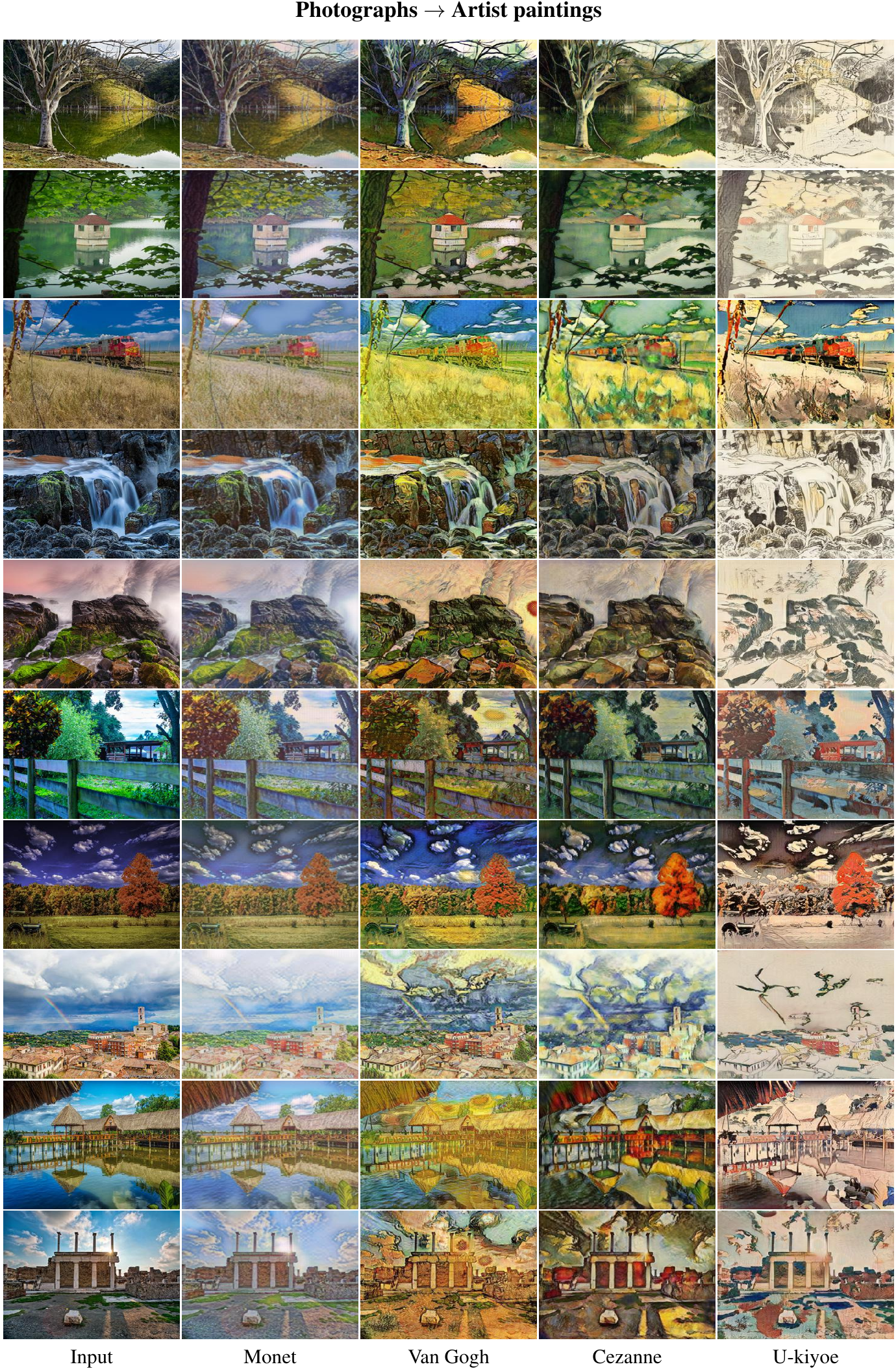}}
  \end{center}
\end{subfigure}%
\captionsetup{font={small}}
\caption{\textbf{Photographs $\to$ Artist paintings.} We translate a photo to the artistic styles of Monet, Van Gogh, Cezanne, and U-kiyoe.}
\label{fig:supp_6}
\end{center}
\end{figure*}

\begin{figure*}[t]
\begin{center}
\begin{subfigure}{0.98\textwidth}
  \begin{center}
  \scalebox{1}[1]{\includegraphics[scale=1.0]{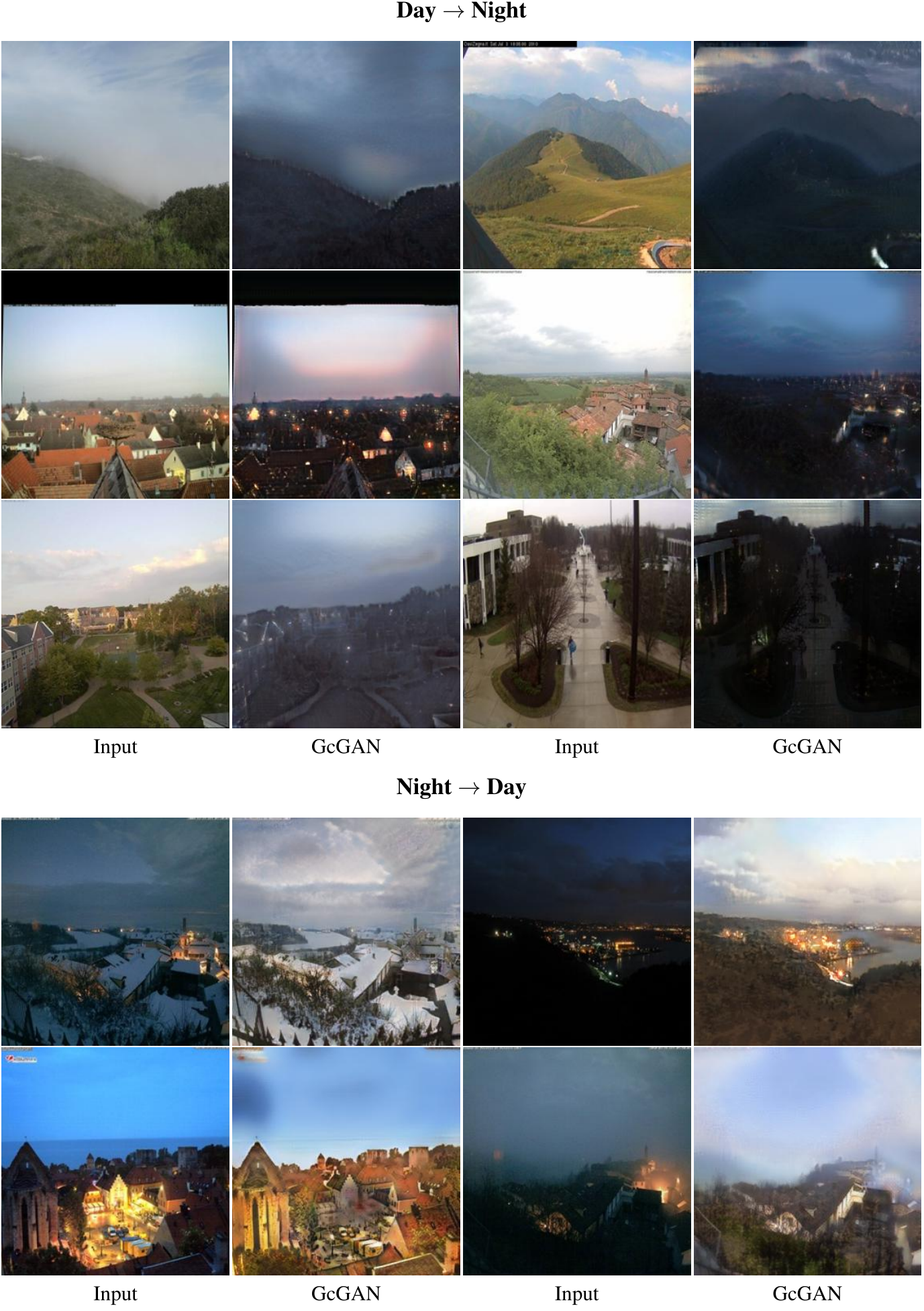}}
  \end{center}
\end{subfigure}%
\captionsetup{font={small}}
\caption{\textbf{Day $\rightleftharpoons$ Night.}}
\label{fig:supp_8}
\end{center}
\end{figure*}

\begin{figure*}[t]
\caption*{\Large{\textbf{Failure Cases}}}
\begin{center}
\begin{subfigure}{0.98\textwidth}
  \begin{center}
  \scalebox{1}[1]{\includegraphics[scale=1.0]{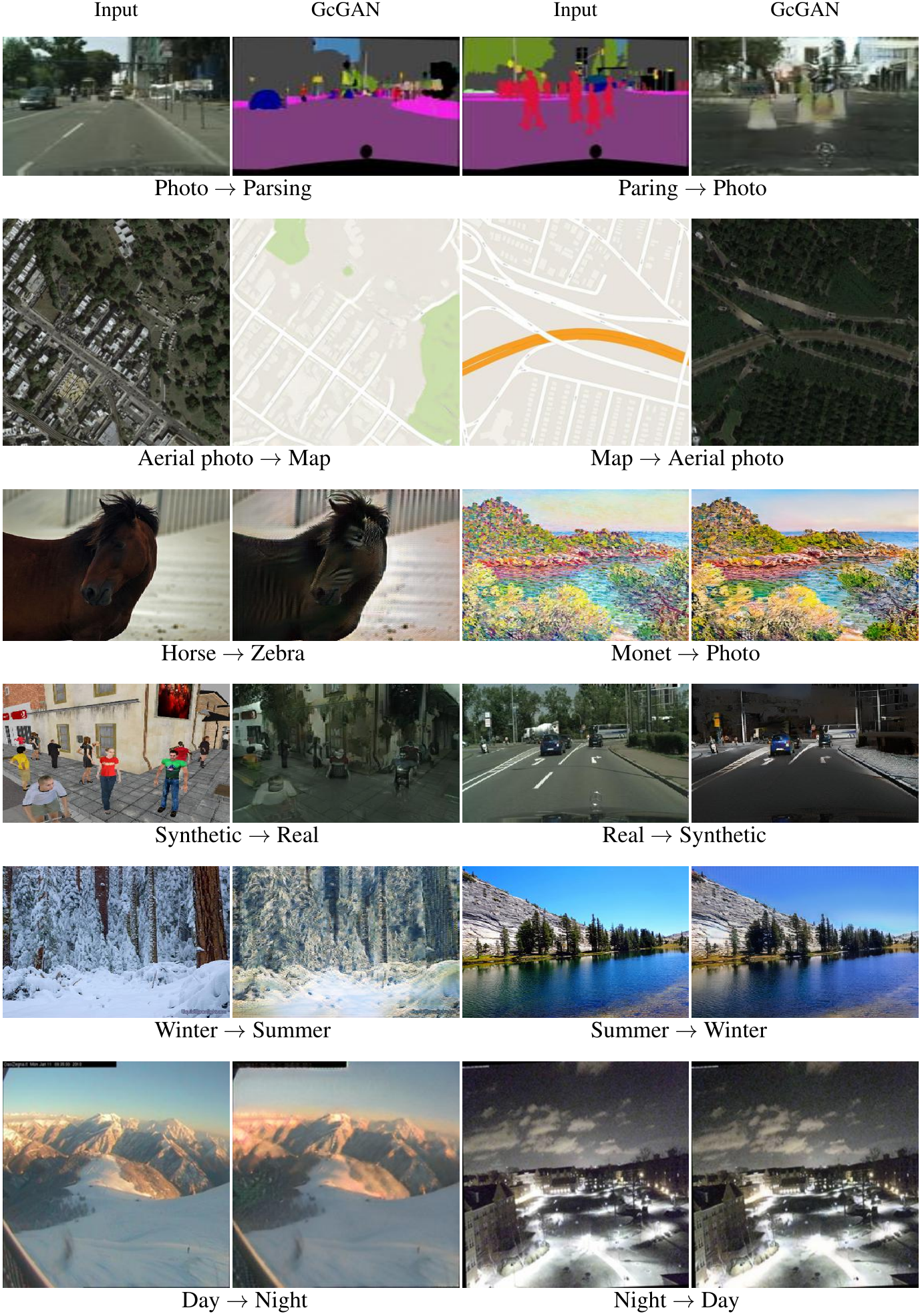}}
  \end{center}
\end{subfigure}%
\captionsetup{font={small}}
\vspace{0.2cm}
\caption{\textbf{Failure Cases.} GcGAN cannot guarantee reasonable translations for all the cases as previous works. Thus, more assumptions and constraints should be investigated to improve unsupervised domain mapping.}
\label{fig:supp_10}
\end{center}
\end{figure*}

\end{document}